\pdfoutput=1

\documentclass[11pt]{article}

\usepackage{EMNLP2023}

\usepackage{times}
\usepackage{latexsym}

\usepackage[T1]{fontenc}

\usepackage[utf8]{inputenc}

\usepackage{microtype}

\usepackage{inconsolata}
\usepackage{graphicx}
\usepackage{tabularx}
\usepackage{adjustbox}
\usepackage{booktabs}
\usepackage{multirow}
\usepackage{amsmath}
\usepackage{bbm}
\usepackage{amssymb}
\usepackage{cuted}
\usepackage{comment}
\usepackage{lipsum}
\usepackage{color}
\usepackage{colortbl}
\usepackage{xcolor}
\usepackage{relsize}
\usepackage{makecell}

\usepackage{subcaption} 

\title{GPT-RE: In-context Learning for Relation Extraction \\ using Large Language Models}

\author{
Zhen Wan$\ \,^1$  \hspace{1em}
Fei Cheng$ \ \,^1$ \hspace{1em} 
Zhuoyuan Mao$ \ \,^1$ \hspace{1em}\\
{\bf Qianying Liu$^1$} \hspace{1em} 
{\bf Haiyue Song$^1$ } \hspace{1em}
{\bf Jiwei Li$^2$ } \hspace{1em} 
{\bf Sadao Kurohashi$^1$ }\\
$^1$ Kyoto University, Japan \hspace{1em}
\\
$^2$ Zhejiang University, China \hspace{1em}
\\
\texttt{\{zhenwan, zhuoyuanmao, ying, song\}@nlp.ist.i.kyoto-u.ac.jp} \\
\texttt{\{feicheng, kuro\}@i.kyoto-u.ac.jp} \\
\texttt{\{jiwei\_li\}@zju.edu.cn} \\
}

\begin{document}
\maketitle
\begin{abstract}

In spite of the potential for ground-breaking achievements offered by large language models (LLMs) (e.g., GPT-3) via in-context learning (ICL), they still lag significantly behind fully-supervised baselines (e.g., fine-tuned BERT) in relation extraction (RE).
This is due to the 
 two major shortcomings of ICL for RE: (1) low relevance regarding entity and relation 
in existing sentence-level demonstration retrieval approaches for ICL; and 
(2) the lack of explaining input-label mappings of demonstrations leading to poor ICL effectiveness. 

In this paper, we propose GPT-RE to 
successfully address the aforementioned issues by (1) incorporating task-aware representations in demonstration retrieval; and (2)
enriching the demonstrations with gold label-induced reasoning logic. 
We evaluate GPT-RE on four widely-used RE datasets and observe that
GPT-RE  achieves improvements over  not only existing GPT-3 baselines, 
but also fully-supervised baselines as in Figure~\ref{fig:performance}.
Specifically, GPT-RE achieves SOTA performances on the Semeval and SciERC datasets, and competitive performances on the 
TACRED and ACE05 datasets. 

Additionally, a critical issue of LLMs revealed by previous work, the strong inclination to wrongly classify \textsc{null} examples into other pre-defined labels, is substantially alleviated by our method. We show an empirical analysis.\footnote{\url{https://github.com/YukinoWan/GPT-RE}}
\end{abstract}

\section{Introduction}
The emergence of large language models (LLMs) such as GPT-3~\cite{NEURIPS2020_1457c0d6,https://doi.org/10.48550/arxiv.2201.08239,DBLP:journals/corr/abs-2204-02311,https://doi.org/10.48550/arxiv.2112.11446,https://doi.org/10.48550/arxiv.2203.15556} 
represents a significant advancement in natural language processing (NLP). 
Instead of following 
a pretraining-and-finetuning pipeline 
~\cite{devlin-etal-2019-bert,beltagy-etal-2019-scibert,https://doi.org/10.48550/arxiv.1910.10683,https://doi.org/10.48550/arxiv.1909.11942,zhuang-etal-2021-robustly}, which finetunes a
pre-trained model on a task-specific dataset in a fully-supervised manner, 
LLMs employ a new paradigm known as in-context learning (ICL) ~\cite{NEURIPS2020_1457c0d6,https://doi.org/10.48550/arxiv.2202.12837}
which formulates an NLP task under the paradigm of language generation and makes predictions by learning from a few demonstrations. 
Under the framework of ICL, LLMs  achieve remarkable performance rivaling previous fully-supervised methods even with only a limited number of demonstrations provided in various tasks such as solving math problems,
commonsense reasoning, text classification, fact retrieval, natural language inference, and semantic parsing~\cite{NEURIPS2020_1457c0d6,DBLP:journals/corr/abs-2202-12837,DBLP:conf/icml/ZhaoWFK021,liu-etal-2022-makes,shin-etal-2021-constrained}.



\begin{figure}[t]
    \centering
    \includegraphics[width=\linewidth]{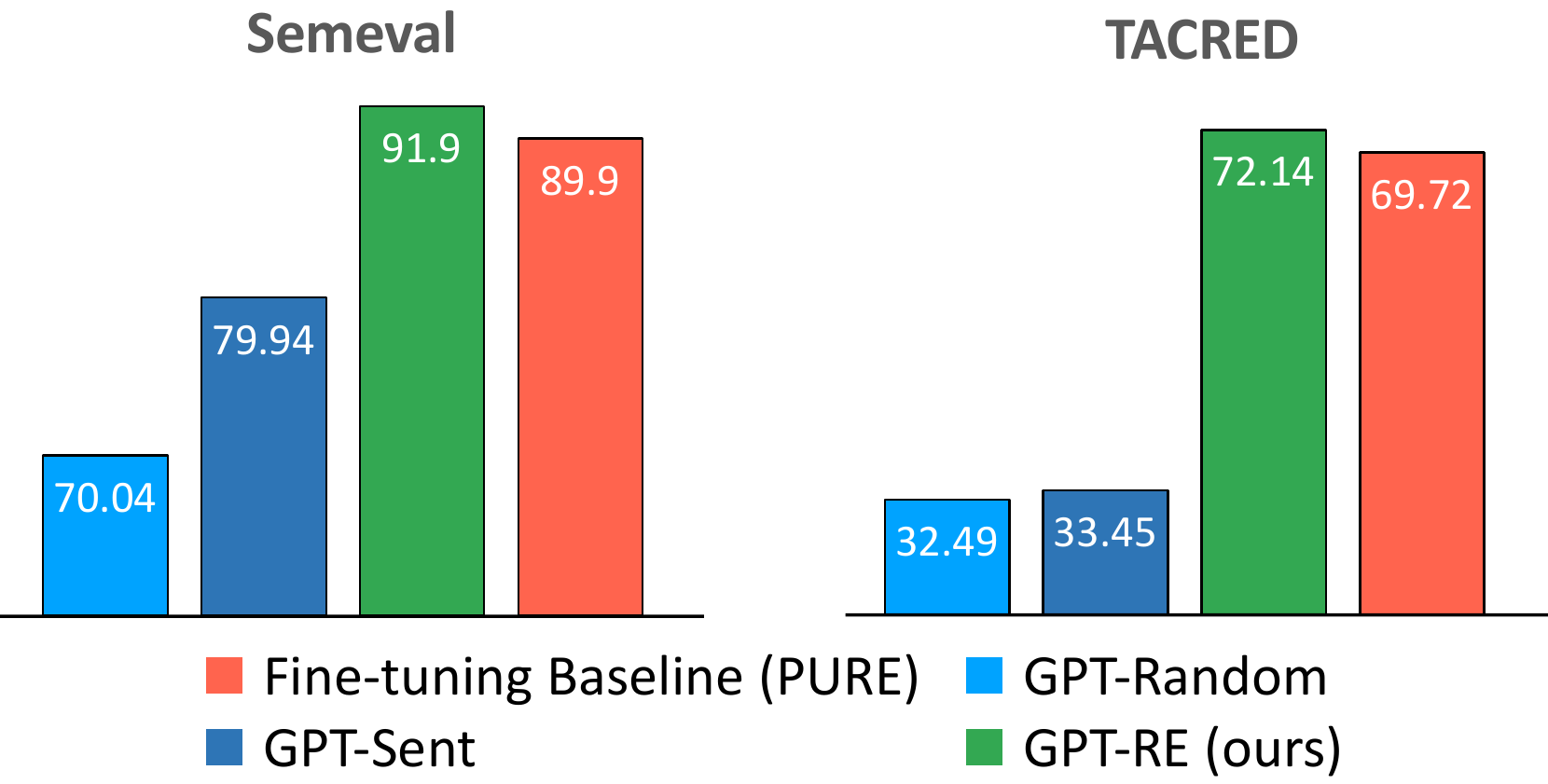}
    \caption{\textbf{Micro F1 performances on two RE datasets}. Previous GPT baselines (\textit{GPT-Random}: randomly selected demonstrations and \textit{GPT-Sent}: sentence-level demonstration retrieval) largely underperform fine-tuning baseline PURE while our \textit{GPT-RE} substantially outperforms all baselines.}
    \label{fig:performance}
\end{figure}


Despite the overall promising performance of LLMs, the utilization of ICL for relation extraction (RE) is still suboptimal.
RE is the central task for knowledge retrieval requiring a deep understanding of natural language, which seeks to identify a pre-defined relation between a specific entity pair mentioned in the input sentence or \textsc{null} if no relation is found. Given a test input, ICL for RE prompts the input of LLMs with the task instruction, a few demonstrations retrieved from the training data, and the test input itself. Then LLMs generate the corresponding relation. 
Recent research ~\cite{DBLP:journals/corr/abs-2203-08410} has sought to apply GPT-3 ICL to biomedical RE, 
but the results 
are relatively negative and 
suggest that GPT-3 ICL still significantly underperforms 
fine-tuned models. 



\begin{figure}[t]
    \centering
    \includegraphics[width=\linewidth]{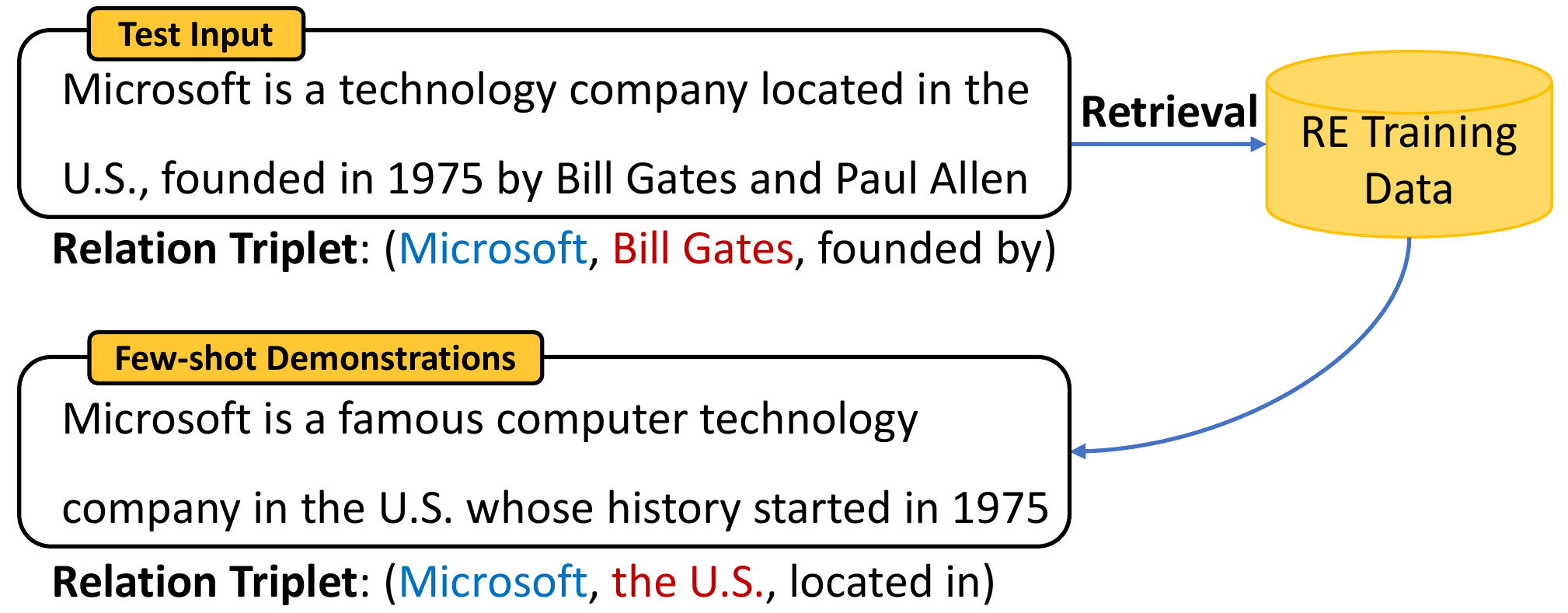}
    \caption{Retrieval without considering the task-aware triplet results in noisy demonstrations. }
    \label{fig:intro_a}
\end{figure}

The reasons that cause the pitfall of GPT-3 ICL in RE are two folds:
(1) The low relevance regarding entity and relation 
in the retrieved demonstrations for ICL.
Demonstrations are selected randomly 
or via $k$-nearest neighbor ($k$NN) search 
based on sentence embedding \cite{liu-etal-2022-makes,DBLP:journals/corr/abs-2203-08410}. 
Regrettably, $k$NN-retrieval based on sentence embedding is more concerned with the relevance of the overall sentence semantics and not as much with the specific entities and relations it contains, which leads to low-quality  demonstrations. 
 As shown in Figure~\ref{fig:intro_a}, the test input retrieves a semantically similar sentence but is not desired in terms of entities and relations.


(2) The lack of explaining input-label mappings in demonstrations leads to poor ICL effectiveness:
A vanilla form of ICL lists all demonstrations as input-label pairs without any explanations. 
This may mislead LLMs to learn shallow clues from surface words, while a relation can be presented in diverse forms due to language complexity. Especially when ICL has a maximal input length, optimizing the learning efficiency of each single demonstration becomes extremely important.

\begin{figure}[t]
    \centering
    \includegraphics[width=\linewidth]{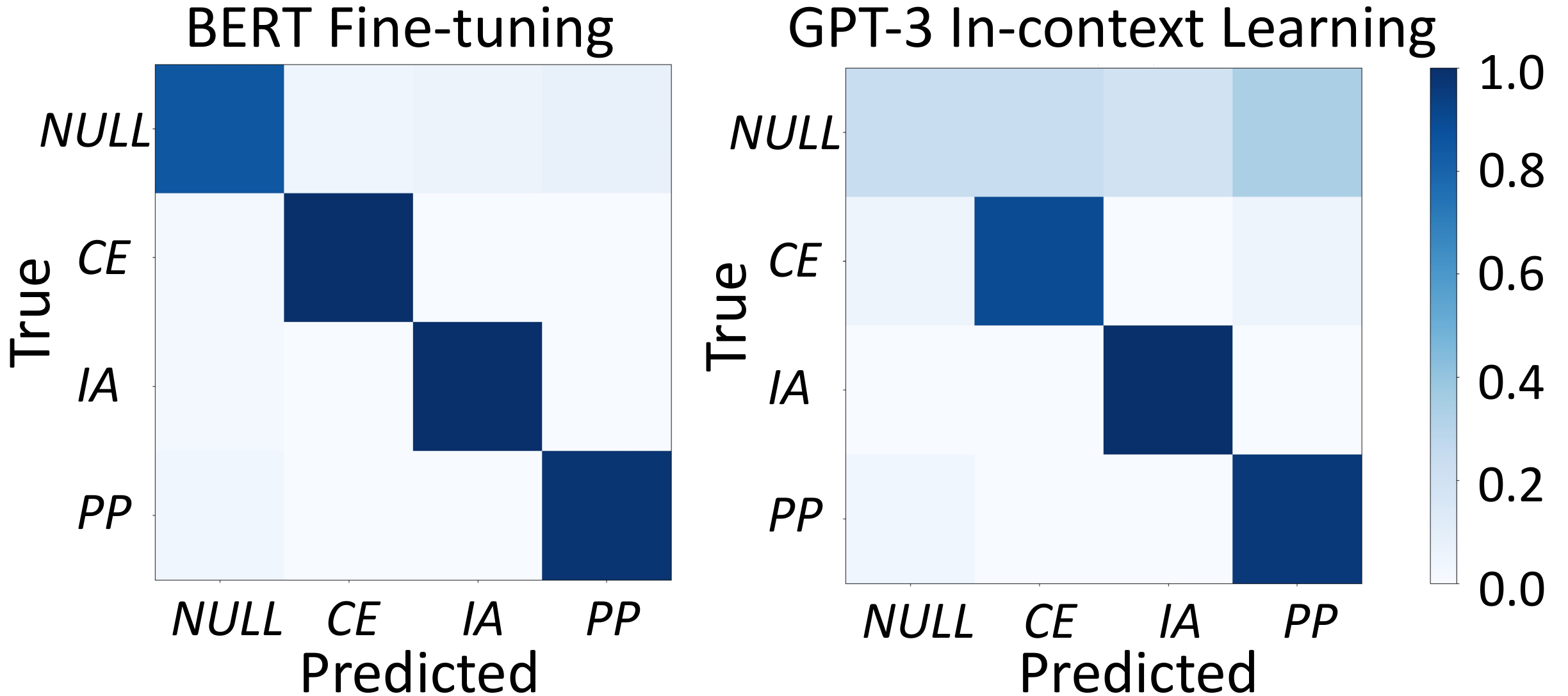}
    \caption{Confusion matrix on Semeval dataset with three selected relation labels. The \textsc{NULL} examples are overpredicted to other relations by GPT-3. CE: Cause-Effect, IA: Instrument-Agency, PP: Product-Producer.}
    \label{fig:intro_b}
\end{figure}


\begin{figure*}[t]
    \centering
    \includegraphics[width=\linewidth]{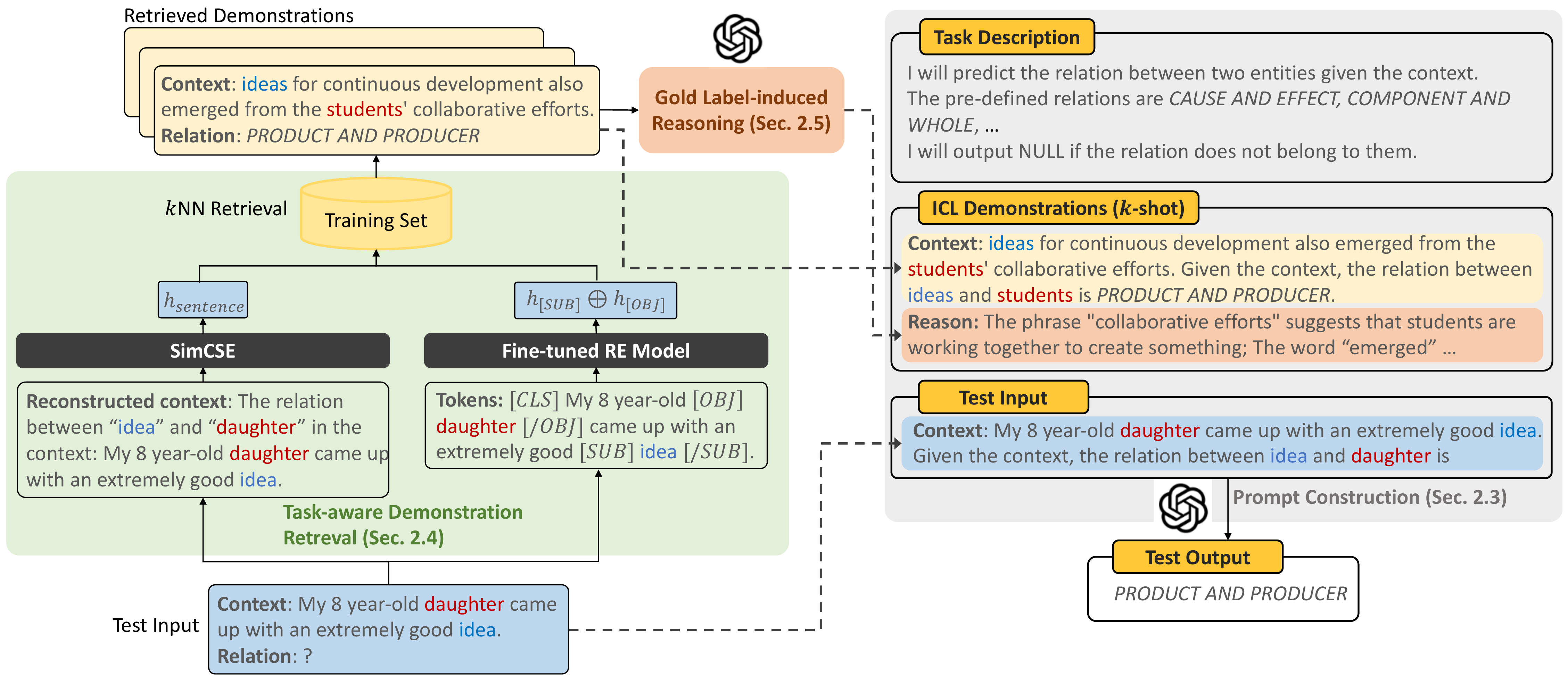}
    \caption{\textbf{An illustration of GPT-RE}. Given a test input, we first leverage two different task-aware retrieval methods to search for highly relevant demonstrations from the training set,
    and then incorporate the gold label-induced reasoning for each demonstration. Above contents will then be included in the prompt construction to make the prediction.} 
    \label{fig:overview}
\end{figure*}

To this end, we propose GPT-RE for the RE task. 
GPT-RE employs two strategies to resolve the issues above: 
(1) \textbf{task-aware retrieval} and (2) \textbf{gold label-induced reasoning}. 
For (1) task-aware retrieval,
its core is to 
use representations that deliberately encode and emphasize entity and relation information rather than sentence embedding
for $k$NN search. 
We achieve this by two different retrieval approaches: (a) entity-prompted sentence embedding; (b) fine-tuned relation representation,
which naturally places emphasis on entities and relations. 
Both methods contain more RE-specific information than sentence semantics, thus effectively addressing the problem of low relevance.

For (2) gold label-induced reasoning, we propose to inject the reasoning logic into the demonstration to provide more evidence to align an input and the label, a strategy akin to the Chain-of-Thought (CoT) research
~\cite{DBLP:journals/corr/abs-2201-11903,DBLP:journals/corr/abs-2203-11171,DBLP:journals/corr/abs-2205-11916}.
But different from previous work, we allow LLMs to elicit the reasoning process to explain not only why a given sentence should be classified under a particular label but also why a \textsc{null} example should not be assigned to any of the pre-defined categories. 
This process significantly improves the ability of LLMs to align the relations with diverse expression forms.



Recent work reveals another crucial problem named ``overpredicting'' as shown in Figure~\ref{fig:intro_b}: 
we observe that LLMs have 
the strong inclination to wrongly classify
\textsc{null} examples into other pre-defined labels
. 
A similar phenomenon has also been observed in other tasks such as NER~\cite{DBLP:journals/corr/abs-2203-08410,DBLP:journals/corr/abs-2211-07830}. 
In this paper, we show that this issue can be alleviated if the representations for retrieval can be supervised with
the whole set of \textsc{null} in the training data.
We evaluate our proposed method on three popular general domain RE datasets: Semeval 2010 task 8, TACRED and ACE05, and one scientific domain dataset SciERC. We observe that
GPT-RE  achieves improvements over not only existing GPT-3 baselines, 
but also fully-supervised baselines. 
Specifically, GPT-RE achieves SOTA performances on the Semeval and SciERC datasets, and competitive performances on the 
TACRED and ACE05 datasets. 


\section{Methodology: GPT-RE}
\subsection{Task Definition}
Let $\mathcal{C}$ denote the input context and $e_{\text{sub}}\in \mathcal{C}$, $e_{\text{obj}}\in \mathcal{C}$ denote the pair of subject and object entity.
Given a set of pre-defined relation classes $\mathbb{R}$, relation extraction aims to predict the relation $y\in \mathbb{R}$ between the pair of entities ($e_{\text{sub}}, e_{\text{obj}}$) within the context $\mathcal{C}$, or if there is no pre-defined relation between them, predict $y=$ \textsc{null}.

\subsection{Overview}
We will first introduce the prompt construction to formalize RE as a language generation task in Sec.~\ref{prompt}. Then to improve the ICL framework for RE, we will introduce two modules: (1) task-aware demonstration retrieval to select higher-quality demonstrations (Sec.~\ref{retrieval module}); (2) gold label-induced reasoning to enrich each demonstration with explanations (Sec.~\ref{reasoning}). In Figure~\ref{fig:overview}, we show the concrete workflow of processing a test input. 

\subsection{Prompt Construction}
\label{prompt}
We construct a prompt for each given test example, which is fed to the GPT-3 model. Each prompt consists of the following components: 

\paragraph{Instructions $\mathcal{I}$}
We provide a succinct overview of the RE task description and the set of pre-defined classes $\mathbb{R}$.
The model is explicitly asked to output the relation, which belongs to the pre-defined classes. Otherwise, the model will output \textsc{null}. 

\paragraph{ICL Demonstrations $\mathcal{D}$} 
We first leverage a task-aware retriever to acquire a $k$-shot demonstration set,
then enrich each demonstration $(x_{i},y_{i})$ with the gold label-induced reasoning $r_{i}$ to build a new set of $(x_{i},y_{i},r_{i})$ as $\mathcal{D}$.

\paragraph{Test Input $x_{test}$} 
Similar to the demonstrations, we offer the test input $x_{test}$, and GPT-3 is expected to generate the corresponding relation $y_{test}$.

In summary, GPT-RE can be formulated as:
\begin{equation}
    p\left(y_{test}\in \mathbb{R}\cup\{\textsc{null}\}| \mathcal{I},\mathcal{D}, x_{test} \right)
\end{equation}


\subsection{Task-aware Demonstration Retrieval}
\label{retrieval module}

Since ICL demonstrations closer to the test sample in the embedding space result in more consistent and robust performance~\cite{liu-etal-2022-makes}.
Recent work~\cite{DBLP:journals/corr/abs-2203-08410,liu-etal-2022-makes} employs the $k$NN to retrieve the most similar examples in the training set as the few-shot demonstrations for each test input. As $k$NN relies on the choice of the embedding space to encode both test input and examples in the training set, they propose to obtain sentence embedding using pre-trained language models, or other improved sentence embedding.




However, using sentence embedding for $k$NN retrieval has a severe drawback: relation extraction focuses on pair-wise entities, which diverge from the semantic meaning of the entire sentence, leading to an ambiguous retrieval using sentence embedding. In this study, we propose two novel methods to provide more robust representations for better retrieval quality: (1) a naive entity-prompted sentence embedding in Sec.~\ref{gtp-re_simcse}; (2) an advanced fine-tuned relation representation in Sec.~\ref{ft_retrieval}.

\subsubsection{Entity-Prompted Sentence Embedding}
\label{gtp-re_simcse}
Given the discrepancy between sentence embedding and relation extraction, the original context is insufficient for demonstration retrieval. Considering the importance of entity information in RE, we propose reconstructing the context by incorporating entity pair information. For example, given the context ``\textit{\underline{He}} has a sister \textit{\underline{Lisa}},'' the reconstructed context with the entity prompted will be ``The relation between `He' and `Lisa' in the context: He has a sister Lisa.'' This approach 
preserves both the semantic meaning of the sentence and the entity pair-centered information during retrieval. In the paper, we employ the latest robust model SimCSE~\cite{gao-etal-2021-simcse} for 
computing sentence embedding-based similarity.
\subsubsection{Fine-tuned Relation Representation}
\label{ft_retrieval}

Compared to prompt entity information into context sentences, a more straightforward solution is to extract the relation representation from a fine-tuned RE model for retrieving demonstrations.

Current BERT-based fine-tuning methods for RE~\cite{baldini-soares-etal-2019-matching,zhong-chen-2021-frustratingly,DBLP:journals/corr/abs-2210-11800} attempts to capture both the context information and the entity information by adding extra marker tokens to highlight the subject and object entities and their types.
Specifically, given an example: ``\textit{\underline{He}} has a sister \textit{\underline{Lisa}}.'', the input tokens are ``\textsc{\small [CLS] [SUB\_PER]} \textit{\underline{He}} \textsc{\small [/SUB\_PER]} has a sister \textsc{\small [OBJ\_PER]} \textit{\underline{Lisa}} \textsc{\small [/OBJ\_PER]}. \textsc{\small [SEP]}'' where ``\textsc{\small PER}'' is the entity type if provided.
Denote the $n$-th hidden representation of the BERT encoder as $\mathbf{h}_n$. Assuming $i$ and $j$ are the indices of two beginning entity markers \textsc{\small [SUB\_PER]} and \textsc{\small [OBJ\_PER]}, we define the relation representation as $\mathbf{Rel}= \mathbf{h}_i \oplus \mathbf{h}_j$ where $\oplus$ stands for concatenation of representations in the first dimension. Subsequently, this representation is fed into a feedforward network for predicting the relation probability $p(y\in \mathbb{R}\cup\{\textsc{null}\}\ |\ \mathbf{Rel})$.

The entity markers have explicitly encoded subject and object entities and the relation representation $\mathbf{Rel}$ is naturally enriched with the entity information. 
We believe this approach can potentially compensate for the limitations of GPT-3 in RE. While GPT-3 ICL has a constraint of limited demonstrations, the fine-tuning process is unbundled and can be done on the whole train data. It has two subsequent merits. First, the relation representations are directly fine-tuned to fit the RE task, which could significantly boost the overall retrieval quality. Second, the overpredicting \textsc{null} issue will be substantially alleviated because the similar \textsc{null} demonstrated can be accurately recognized by the fine-tuned model.

\begin{figure}[t]
    \centering
    \includegraphics[width=\linewidth]{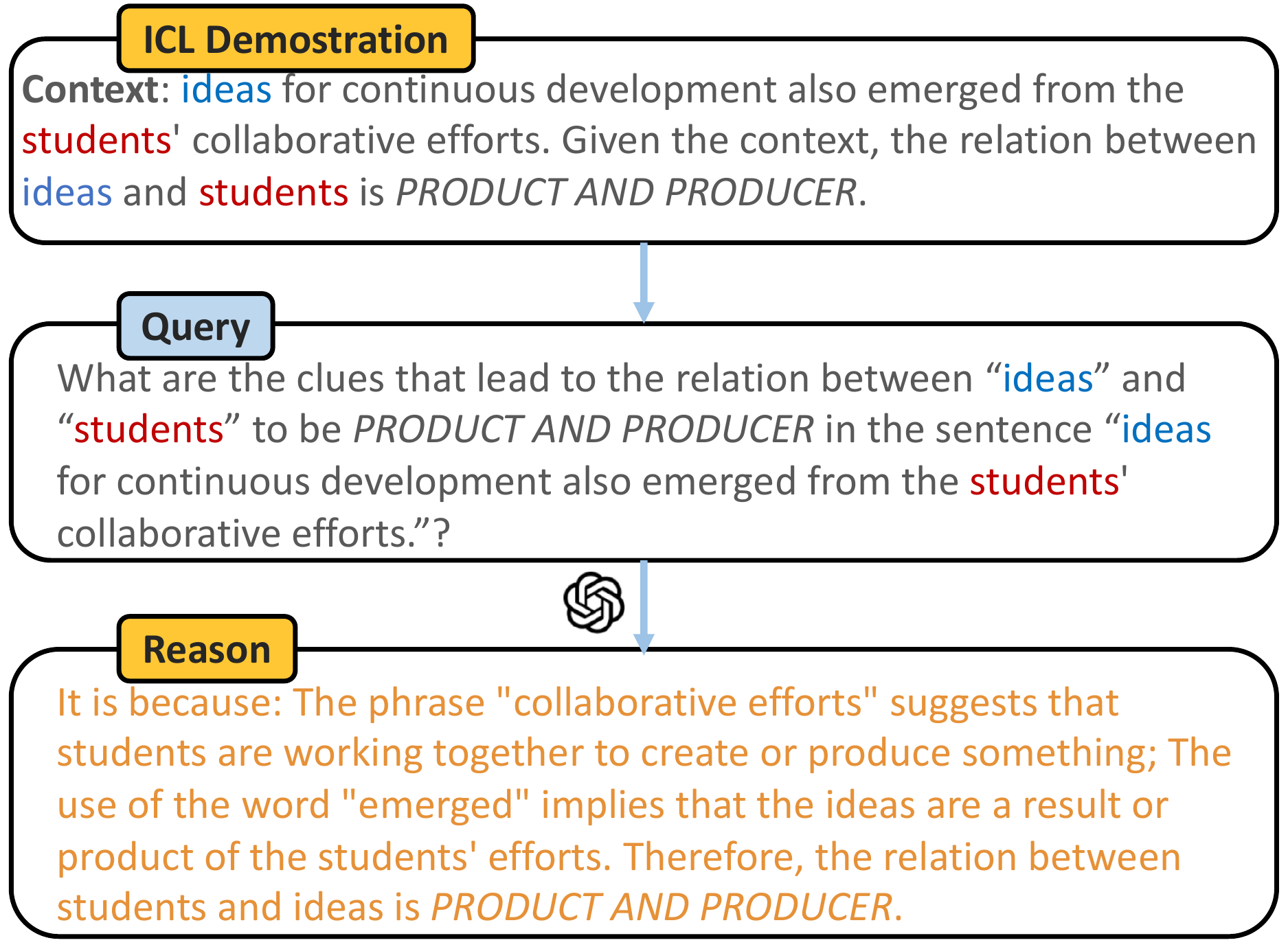}
    \caption{\textbf{An illustration of adding reasoning}. }
    \label{fig:reason}
\end{figure}

\begin{table}
    \centering
    \resizebox{1.0\linewidth}{!}{
    \begin{tabular}{lrrrrr}
    \toprule
        Dataset & \# Relation & \# Train & \# Dev & \# Test (\# Subset) & NULL (\%) \\
        \hline
        Semeval & 9 & 6,507 & 1,493 & 2,717 (2,717) &17.40\%\\
        TACRED & 41 & 68,124 & 22,631 & 15,509 (1,600) &79.40\% \\
        SciERC & 7 & 16,872 & 2,033 & 4,088 (4,088) &90.16\%\\
        ACE05 & 6 &121,368 & 27,597 & 24,420 (2,442)& 95.60\%\\
        \bottomrule
    \end{tabular}
    }
    \caption{\textbf{Statistics of datasets}.}
    \label{stat}
\end{table}
\subsection{Gold Label-induced Reasoning}
\label{reasoning}
Recent CoT work has reported significant progress in the commonsense and numerical reasoning tasks by automatically eliciting the reasoning steps for solving a question. 
While in the RE task, two entities can possibly hold multiple relations, e.g., ``Joe Biden'' can be either the president of or lives in ``U.S.''. The reasoning generation could be out of focus if it lacks interaction with the gold label. 

\begin{table*}[t]
    \centering
    \begin{tabular}{llrrrr}
    \toprule
        Methods &Retriever& \;\;\;\;Semeval\;\;\;\, &  \;\;\;\;TACRED\;\;\;\, & \;\;\; SciERC\;\;\;\, & \;\;\;ACE05\;\;\;\,\\

         \toprule
         \multicolumn{6}{c}{\textit{GPT-3 Baselines (Best $k$-shot)}} \\
         \hline
         GPT-Random &-& 70.04 \small($30$)  & 32.49 \small($15$) & 17.92 \small($25$) & 9.04 \small($25$)\\
         GPT-Sent &SimCSE& 79.94 \small($30$)  & 33.45 \small($15$) &20.96 \small($25$) & 6.31 \small($25$) \\
         \bottomrule
         \multicolumn{6}{c}{\textit{Ours (Best $k$-shot)}} \\
         \hline
        
         GPT-RE\_SimCSE &SimCSE & 81.02 \small($30$)  & 37.44 \small($15$) & 26.46 \small($25$)& 8.67 \small($25$) \\
         GPT-RE\_SimCSE* &SimCSE & 77.49 \small($15$)   & 31.58 \small($10$) & - & - \\
         \ \ + Reasoning &SimCSE& 79.88 \small($15$) & 33.18 \small($10$)  & - & - \\
         \hline
         
         GPT-RE\_FT &PURE & \textbf{\underline{91.90}} \small($25$) & \underline{72.14} \small($15$)& \textbf{\underline{69.00}} \small($30$)& 68.73 \small($25$)\\
         GPT-RE\_FT* &PURE & \underline{91.11} \small($15$)  & \underline{70.38} \small($10$) & -& - \\
         \ \ + Reasoning&PURE& \underline{91.82} \small($15$) & \underline{70.97} \small($10$) & -& -\\
         \toprule
         \multicolumn{6}{c}{\textit{Fine-tuned RE Baselines}} \\
         \hline
         ~\citet{DBLP:journals/corr/abs-2010-04829} &&\textbf{91.90}\;\;\;\;\;\, & -\;\;\;\;\;\, & -\;\;\;\;\;\, &  -\;\;\;\;\;\,\\
         ~\citet{wang-etal-2022-deepstruct} &&-\;\;\;\;\;\,  & $\clubsuit$\textbf{76.80}\;\;\;\;\;\, &  -\;\;\;\;\;\,&-\;\;\;\;\;\, \\
         \multicolumn{2}{l}{PURE~\cite{zhong-chen-2021-frustratingly}}&89.90\;\;\;\;\;\,& 69.72\;\;\;\;\;\, & 68.45\;\;\;\;\;\, &\textbf{70.09}\;\;\;\;\;\, \\
         \bottomrule
    \end{tabular}
    \caption{\textbf{Main Results on four RE datasets}. All results are given by Micro-F1. * denotes the same $k$-shot for the comparison with + Reasoning. Due to the costly GPT-3 expense, we conducted Reasoning experiments on  the two relatively smaller datasets Semeval and TACRED. $\clubsuit$ denotes that this performance is not comparable as it evaluates on the entire test set. The \underline{underline} denotes the results outperforming the fine-tuning baseline PURE.} 
    \label{table: main results}
\end{table*}
 
In this section, we propose to let GPT-3 induce the reasoning logic for each demonstration by the corresponding gold relation label.
As shown in Figure~\ref{fig:reason}, given a selected demonstration, we first generate a query prompt ``What are the clues that lead to the relation between [entity1] and [entity2] to be [relation] in the sentence [context]?'' based on the demonstration and subsequently ask GPT-3 to generate clues ``It is because: ...'' on the labeled relation between the pair of entities in the context. Finally, we augment the demonstration by incorporating the generated clues induced by GPT-3.
\section{Experiment Setup}

\subsection{Datasets}
We evaluate on three popular general domain RE datasets and one scientific domain dataset. Due to the cost of running the model in the API with GPT-3, in our main results, we sample a subset (See Appendix~\ref{subset}) from the original test set for two datasets: ACE05 and TACRED as shown in Table~\ref{stat}.

\paragraph{Semeval 2010 task 8}~\citet{hendrickx-etal-2010-semeval} focuses on semantic relations between pairs of nominals collected from general domain resources.
\paragraph{TACRED}~\citet{zhang-etal-2017-position} is a large-scale relation extraction dataset with 106,264 examples built over newswire and web text. 

\paragraph{SciERC}~\citet{luan-etal-2018-multi} collects AI paper abstracts and annotated relations, especially for scientific knowledge graph construction.

\paragraph{ACE05} contains the entity, relation, and event annotations collected from domains including newswire,
broadcast, discussion forums, etc.

\subsection{Baseline Methods}
\paragraph{GPT-3 baselines}
For GPT-3 baselines and our methods, we select ``\texttt{text-davinci-003}'' with maximal 4,097 input tokens and use the identical prompt construction (Sec.~\ref{prompt}) via OpenAI API. 
We implement two
categories of GPT-3 baselines:

\noindent\textbf{(1) GPT-Random}
Instead of randomly selecting few-shot demonstrations from the training data for each test input, we add extra constraints to make the label distribution of selected demonstrations more uniform. Our preliminary experiments suggest that this is a stronger baseline than the vanilla random.

\noindent\textbf{(2) GPT-Sent}
Previous work attempts various sentence embedding in retrieval.
In this work, our implementation adopted SimCSE~\cite{gao-etal-2021-simcse}, which has been demonstrated to be the state-of-the-art method for sentence similarity tasks. 

\paragraph{Fine-tuned RE Models}
In our experiment, we choose PURE~\cite{zhong-chen-2021-frustratingly}, an entity marker-based fine-tuned model mentioned in Sec.~\ref{ft_retrieval} to obtain the representations for retrieval. Meanwhile, PURE performs as a directly comparable baseline.
We also compare with corresponding SOTA fine-tuned baselines on Semeval~\citet{DBLP:journals/corr/abs-2010-04829} (reformulate RE as the question answering task) and TACRED~\citet{wang-etal-2022-deepstruct} (extra pre-training to capture RE structure) datasets.

All implementation details are in Appendix~\ref{hyperparameters}.

\begin{figure}[t]
    \centering
      \begin{subfigure}[b]{0.49\textwidth}
         \centering
         \includegraphics[width=\textwidth]{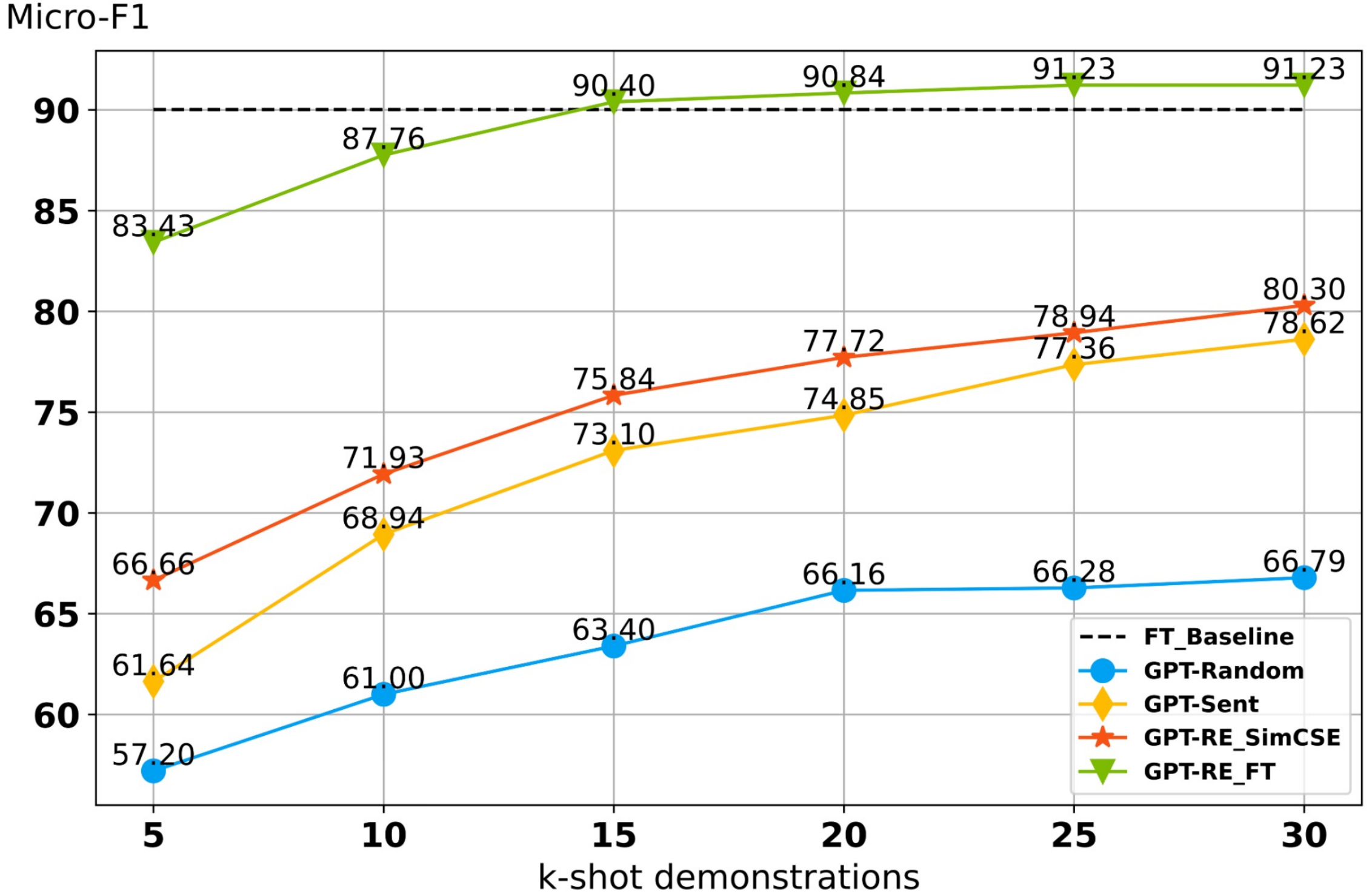}
         \caption{\textbf{The comparison on retrieval modules}}
         \label{result:k_shot}
     \end{subfigure}
     \begin{subfigure}[b]{0.49\textwidth}
         \centering
         \includegraphics[width=\textwidth]{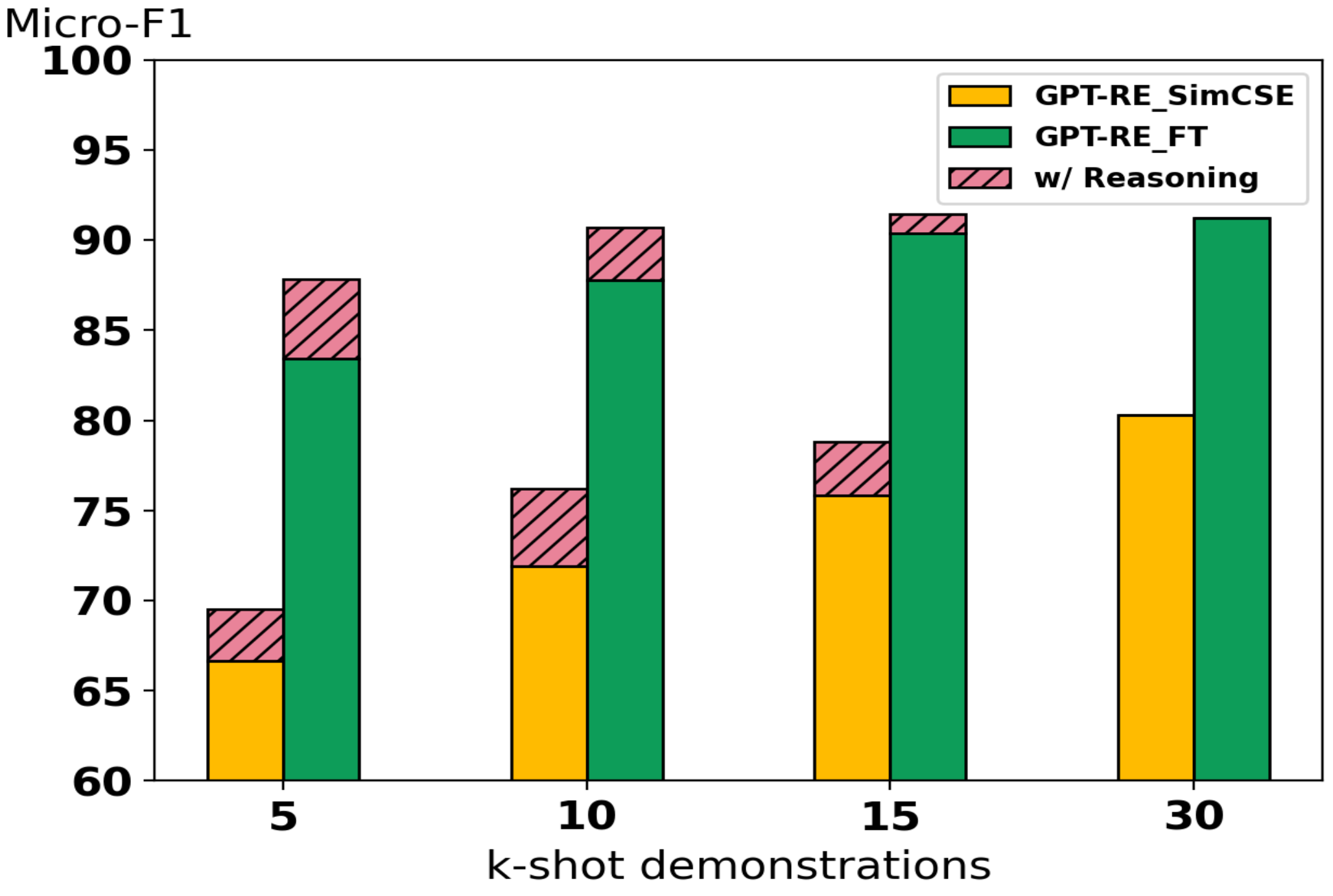}
         \caption{\textbf{Reasoning with fewer demonstrations}.}
         \label{result:reason}
     \end{subfigure}    
    \caption{\textbf{Ablation study on the retrieval and reasoning components on Semeval.} We sampled a subset from the test data with 300 examples. We show the `w/o reasoning' results with $k=30$ for comparison.}
    
    \label{fig:k-shot and reason}
\end{figure}

\section{Experimental Results}
\subsection{Main Results}
We compare our main experiment results with previous methods in Table~\ref{table: main results}. \textbf{GPT-RE\_SimCSE} denotes our entity-prompted sentence embedding for retrieval and \textbf{GPT-RE\_FT} denotes our fine-tuned relation representation for retrieval.
From the table, we can observe that: (1) both \textit{GPT-RE\_SimCSE} and \textit{GPT-RE\_FT} outperform the retrieval-based \textit{GPT-Sent}, indicating that it is necessary to inject the task-specific information into sentence embedding for selecting proper demonstrations; (2) 
\textit{GPT-RE\_FT} succeeds to outperform the fine-tuning baseline PURE on three datasets by $+2.00$, $+2.42$, $+0.55$ Micro-F1. It suggests that GPT-3 has the potential to beat fine-tuning when the retriever has prior task knowledge. \textit{GPT-RE\_FT} eventually achieves SOTA results on Semeval and SciERC. (3) reasoning module improves \textit{GPT-RE\_SimCSE} by around $2\%$ Micro-F1, indicating that gold label-induced reasoning successfully enriches the knowledge of demonstrations. Meanwhile, the high-quality demonstrations obtained by \textit{GPT-RE\_FT} offset the effort of enriching reasoning into demonstrations, which shows relatively trivial improvements. 
Since reasoning aims at enriching demonstrations, this feature potentially works better with fewer demonstrations, as shown in Section~\ref{ab:reasoning}.

\begin{figure}[t]
    \centering
    \includegraphics[width=1.0\linewidth]{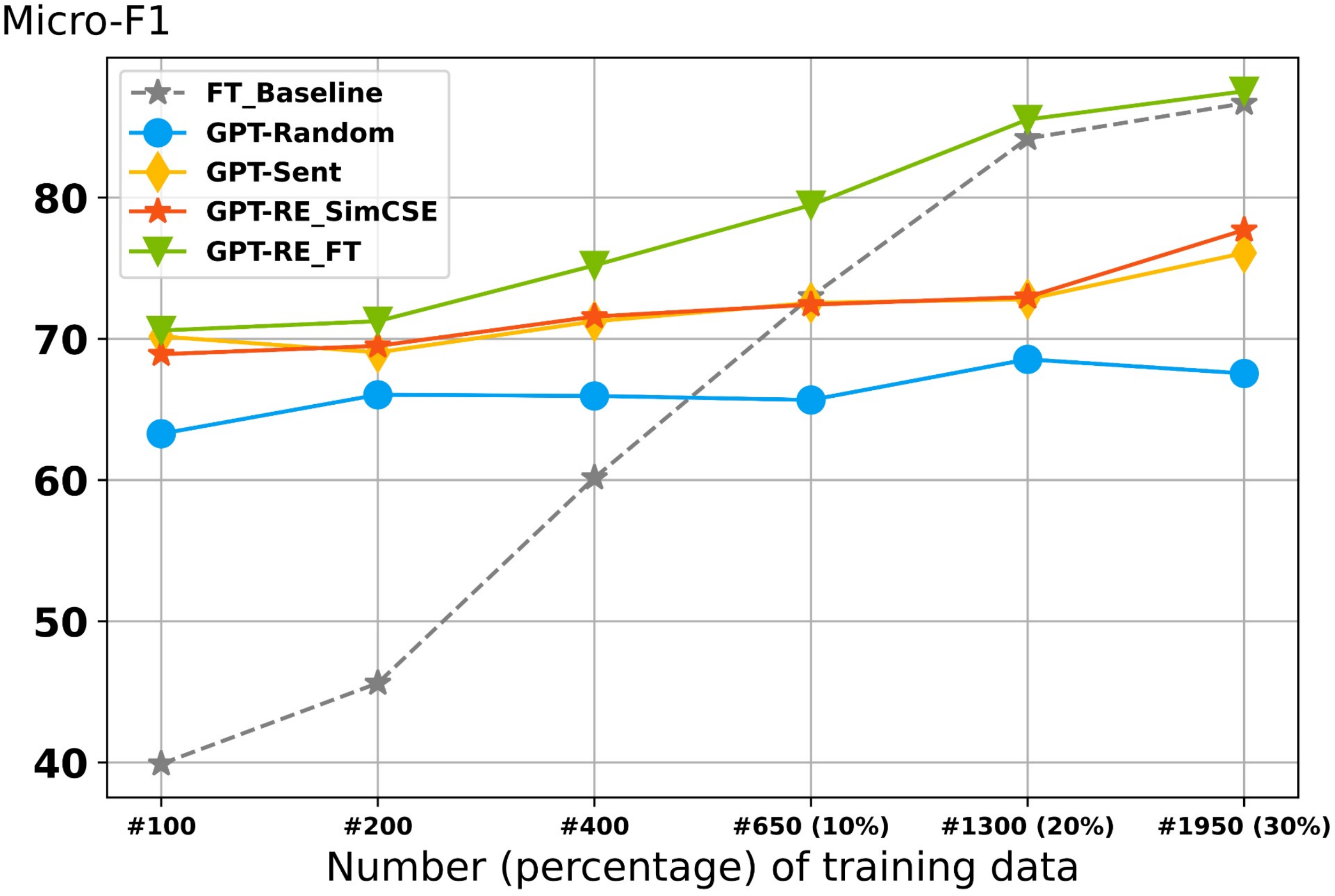}
    \caption{\textbf{Low-resource Scenario on Semeval}. We limit the percentage of training data for both fine-tuning and retrieval in GPT-RE. 
    }
    \label{result:low-resource}
\end{figure}

\subsection{Ablation Study on Task-aware Retrieval}
We first implement the ablation experiments of the retrieval component with the setting of increasing $k$-shot demonstrations (Figure~\ref{result:k_shot}). We find that: (1) compared to \textit{GPT-Random}, all the retrieval-based models have higher F1 scores and large gradients of the performance curves. It means that GPT-3 can learn from high-quality demonstrations more effectively; 
(2) after adding entity information to the SimCSE retrieval, \textit{GPT-RE\_SimCSE} achieves better performance throughout all $K$ shots, indicating that task-aware sentence embedding can capture the feature of RE and provide more proper demonstrations; (3) finally, the fine-tuned relation representation retriever \textit{GPT-RE\_FT} significantly outperforms all retrieval-based methods and beats the fine-tuning baseline when $k>15$. Note that even with $k=5$ demonstrations, \textit{GPT-RE\_FT} still works better than \textit{GPT-RE\_SimCSE} with $k=30$ ($80.30\xrightarrow{} 83.43 (+3.13)$), which indicates that the quality of demonstrations shows much more important than the number of demonstrations.

\begin{figure*}[t]
    \centering
      \begin{subfigure}[b]{0.49\textwidth}
         \centering
         \includegraphics[width=\textwidth]{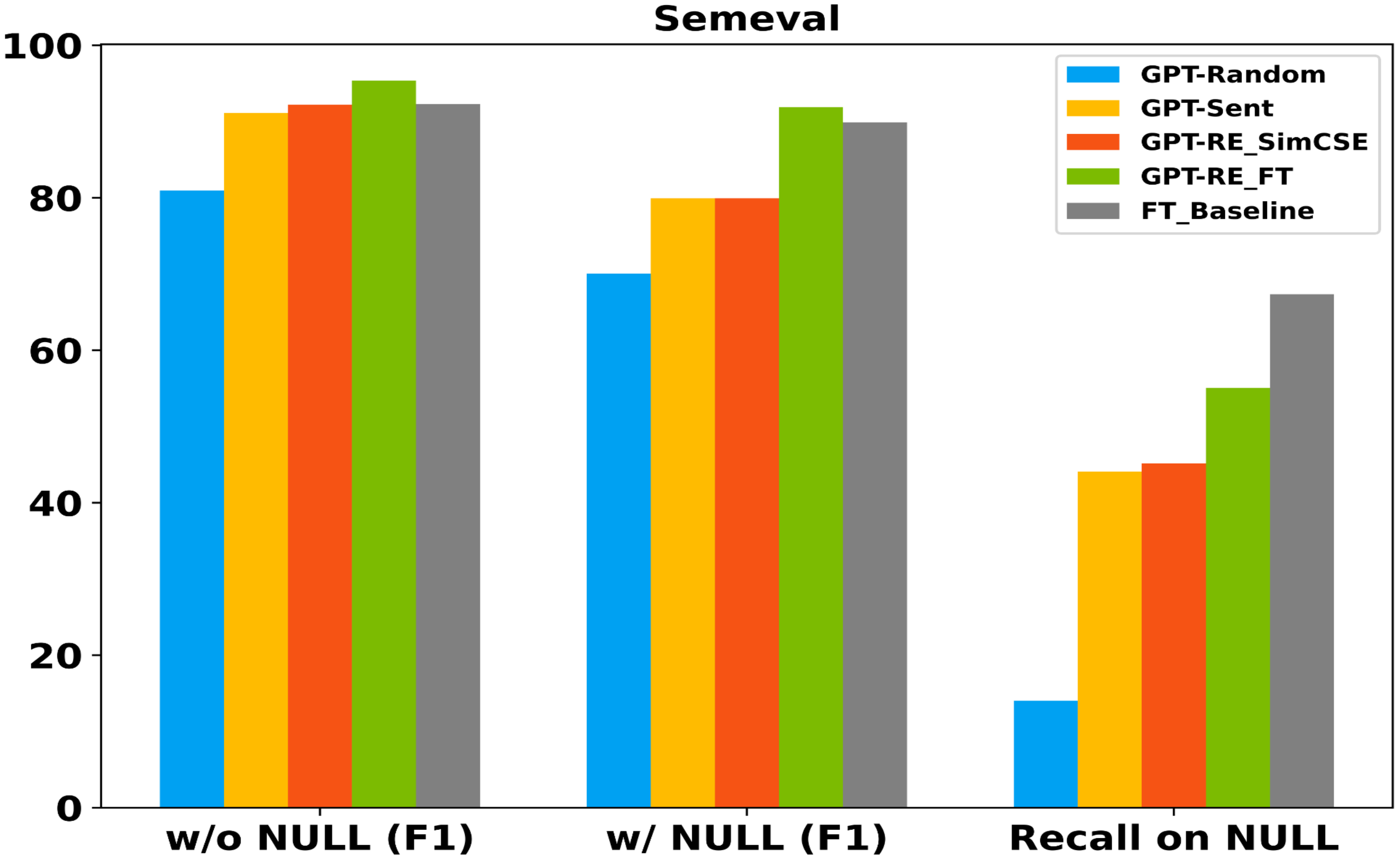}
         \caption{General domain with $17.4\%$ \textsc{null} examples}
         \label{fig:na_semeval}
     \end{subfigure}
     \begin{subfigure}[b]{0.49\textwidth}
         \centering
         \includegraphics[width=\textwidth]{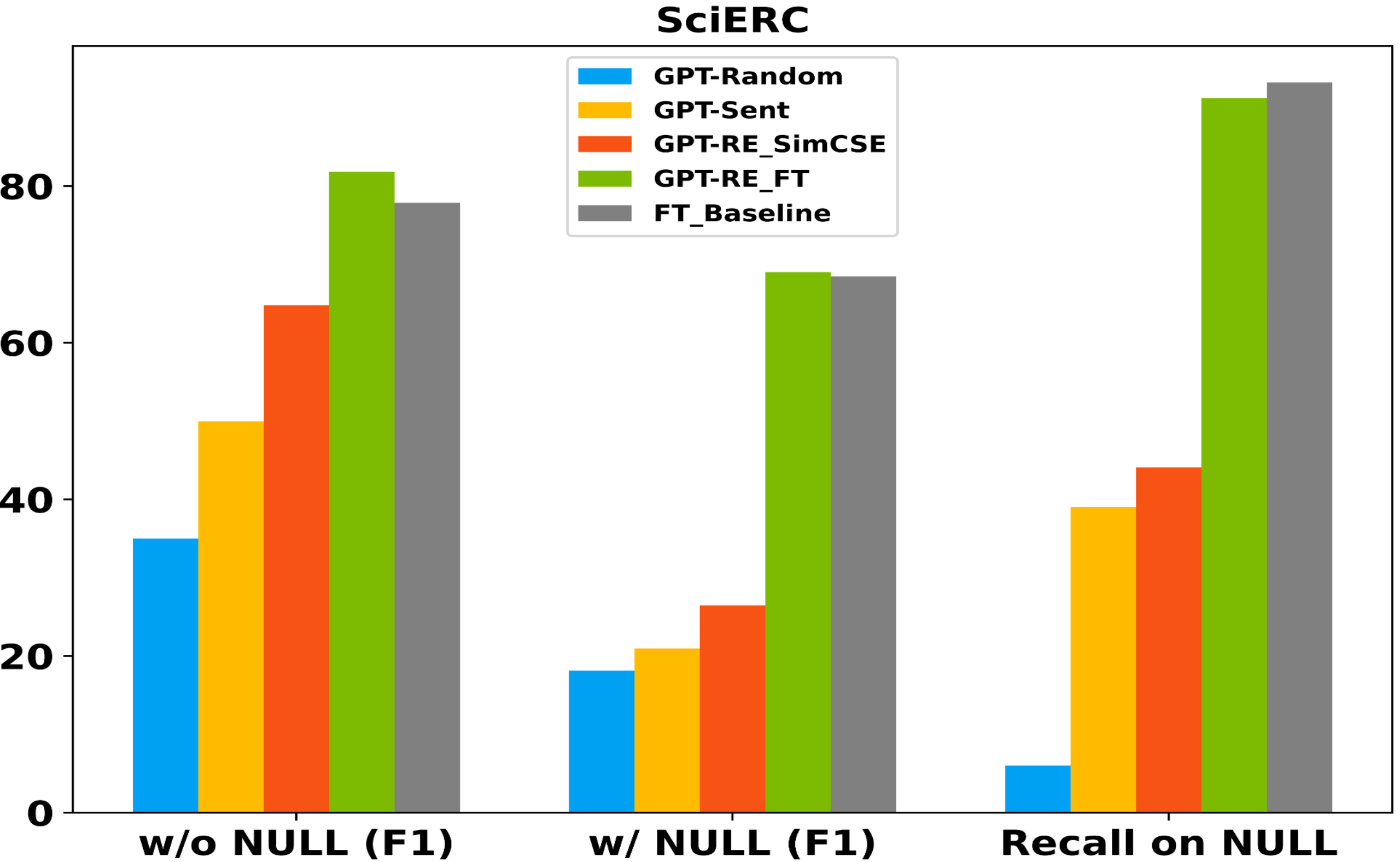}
         \caption{Scientific domain with $90.16\%$ \textsc{null} examples}
         \label{fig:na_scierc}
     \end{subfigure}    
    \caption{\textbf{Analysis on the effects of \textsc{null} examples}. \texttt{w/o} \textsc{null} refers to the classification setting that \textsc{null} examples are excluded from the train and test data. \texttt{w/} \textsc{null} refers to the original extraction setting. We use the full test set for the evaluation.}
    
    \label{fig:na_ab}
\end{figure*}

\subsection{Ablation Study on Reasoning Enhancing}\label{ab:reasoning}
We then check the influence of our proposed reasoning-enhanced demonstration, as shown in Figure~\ref{result:reason}. Due to the limited amount of input tokens of GPT-3, we have to set the $k\leq15$ for the tokens of reasoning, leading to a trade-off between adding reasoning and adding more demonstrations. From the result, we find that: (1) with reasoning-enhanced demonstrations, GPT-3 always achieves better scores across all the $k$-shot settings of both \textit{GPT-RE\_SimCSE} and \textit{GPT-RE\_FT}, indicating that the reasoning induced from ground truth relation labels can effectively unlock the reasoning ability of GPT-3 and improve the ICL with a deeper understanding of demonstrations. Specifically, for \textit{GPT-RE\_FT}, the performance improvement becomes less significant when more demonstrations are provided, which is feasible as with more high-quality demonstrations available, GPT-3 can already learn the internal reasoning behind each demonstration; (2) since the reasoning enhancement works better with fewer demonstrations, we expect this method can be an effective solution to low-shot relation extraction~\cite{han-etal-2018-fewrel,DBLP:conf/cikm/GengCZSZ20,liu-etal-2022-pre}, which aims at recognizing novel relations with very few or no examples, and we leave this for future work.

\subsection{Low-resource Scenario}
We conduct the experiment for observing the low-resource performance in the general domain Semeval task.
As shown in Figure~\ref{result:low-resource}, we observe that: (1) all the GPT-3 based results work better than fine-tuning in when the training examples are less than \# $650$ ($10\%$). It indicates that in the general domain RE, GPT-3 benefits from its abundant prior knowledge to understand the relations; (2) \textit{GPT-RE\_SimCSE} starts to show a substantial difference to \textit{GPT-Sent} after the training size surpasses 30\%. We believe fewer training candidates could limit the effects of retrieval;
(3) \textit{GPT-RE\_FT} achieves an upper bound performance in all settings, even when the fine-tuned model shows poor performance with hundreds of training data (from \#$100$ to \#$400$). This emphasizes the impressive effectiveness of fine-tuned relation representations for capturing higher-quality demonstrations.
The observation in the low-resource setting is very different from~\newcite{DBLP:journals/corr/abs-2203-08410}. We assume the difference could be caused by the domain and \textsc{null} proportion of the task.

\section{Analysis}
\subsection{The Issue of ``Overpredicting''}
To analyze the influence of \textsc{null} class, we compare the effectiveness of each method for alleviating this issue on two datasets: general domain Semeval with $17.4\%$ \textsc{null} examples and scientific domain SciERC with $90.16\%$ \textsc{null} examples. As shown in Figure ~\ref{fig:na_ab}, (1) by comparing the performance on Semeval and SciERC, a larger percentage of \textsc{null} examples results in more significant performance drop showing the negative influence of overpredicting \textsc{null} examples; (2) by comparing \texttt{w/o} \textsc{null} and \texttt{w/} \textsc{null}, our \textit{GPT-RE\_FT} shows the most robustness to the influence of \textsc{null} examples, indicating that the RE fine-tuned representations in retrieval can release the overpredicting issue of GPT-3 by providing higher-quality demonstrations; (3) however, even with task-aware representations, all GPT-3 methods still underperform the fine-tuning baseline on \textsc{null} examples, this is due to the confusing definition of \textsc{null}, in many cases, there is a certain relation between entities in the context, but out of the distribution of pre-defined classes. In these cases, GPT-3 tends to overpredict as the relation information may be covered in its prior knowledge. We think this ability of GPT-3 can be useful in more open fields, such as open RE ~\cite{banko-etzioni-2008-tradeoffs} which has no pre-defined relation classes.
\subsection{Case Study of Demonstration Quality}
\begin{figure}[t]
    \centering
    \includegraphics[width=1.0\linewidth]{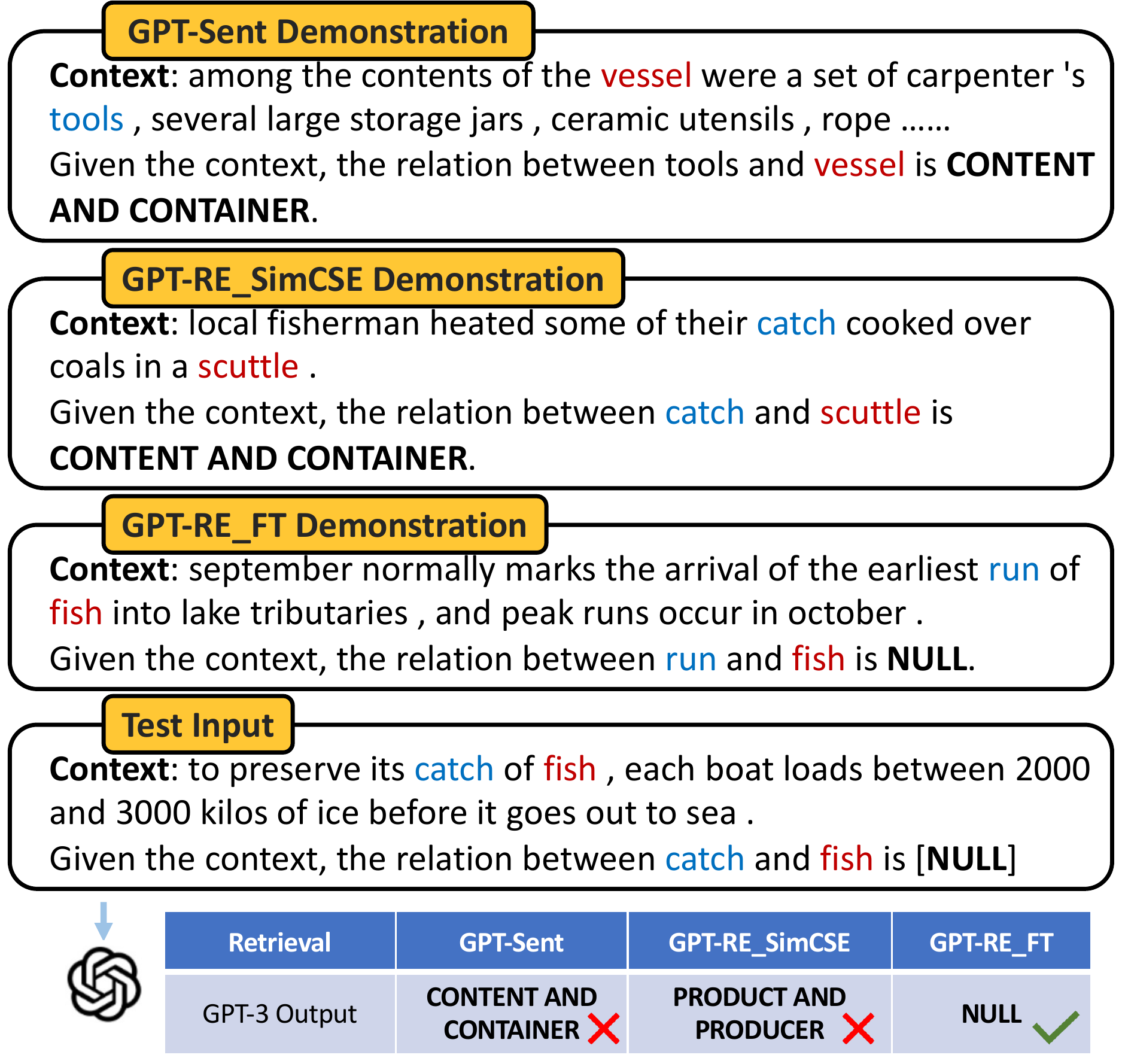}
    \caption{\textbf{A case study of demonstration quality on Semeval.} [\textsc{null}] is the gold label here. }
    \label{case study}
\end{figure}

We select one typical test example to better illustrate the amendment of our task-aware demonstration retrieval. As shown in Figure~\ref{case study}, given the \textsc{null} Example, we show the most similar demonstration in retrieval based on three methods. The \textit{GPT-Sent} retrieved demonstration focuses on the semantic meaning of ``CONTENT AND CONTAINER'' which is shared in the test context, but not revealed in the target entity pair. This mismatch confirms the problem of lacking entity information in retrieval. Instead, \textit{GPT-RE\_SimCSE} retrieves a much more relevant demonstration that shows the same semantic relation between ``catch'' and ``fish'' but still faces a minor mismatch as the gold label is between ``catch'' and ``scuttle.'' Finally, \textit{GPT-RE\_FT} demonstration shares a similar structure with the test input regarding the pair of entities, which is the key clue for predicting the relation between entities. This result shows a level-by-level enhancement with more entity information provided in retrieval. We also show some other case examples in Appendix~\ref{case}.
\section{Related Work}
\paragraph{In-context Learning}
Recent work shows that ICL of GPT-3~\cite{NEURIPS2020_1457c0d6} can perform numerous tasks when provided a few examples in a natural language prompt. 
Existing work focuses on various aspects to effectively utilize the advantages of GPT-3, from prompt design ~\cite{DBLP:conf/nips/PerezKC21} for proper input to coherence calibration ~\cite{malkin-etal-2022-coherence} for tackling the diverse generated output. Another research path locates in the demonstration part, including ordered prompts ~\cite{lu-etal-2022-fantastically} and retrieval-based demonstrations~\cite{rubin-etal-2022-learning,liu-etal-2022-makes,shin-etal-2021-constrained}.

To the best of our knowledge, there is no previous work exploring the potential of GPT-3 on general domain RE tasks. A recent work attempts to leverage GPT-3 in biomedical information extraction (NER and RE), and reveals issues of ICL that may be detrimental to IE tasks in general. Our work succeeds in overcoming these issues to some extent and confirms the potential of GPT-3 in both general and the scientific domain RE.

\paragraph{Retrieval-based Demonstrations}
Several studies have demonstrated that dynamically selecting few-shot demonstrations for each test example, instead of utilizing a fixed set, leads to significant improvement in GPT-3 ICL ~\cite{liu-etal-2022-makes, shin-etal-2021-constrained,rubin-etal-2022-learning}.
They also show that nearest neighbor in-context examples yield much better results than the farthest ones. This leads to the significance of better retrieval modules for demonstrations. Existing attempts rely on sentence embedding in retrieval, including the sentence encoders of PLMs such as BERT~\cite{devlin-etal-2019-bert}, RoBERTa~\cite{zhuang-etal-2021-robustly}
KATE~\cite{liu-etal-2022-makes}
, SimCSE~\cite{gao-etal-2021-simcse}, Sentence-BERT \cite{reimers-gurevych-2019-sentence,wolf-etal-2020-transformers}. Unlike these sentence embeddings, we propose to fine-tune PLMs on our target RE tasks to produce more task-specific and robust representations for retrieval.
\section{Conclusions}


This work explores the potential of GPT-3 ICL on RE for bridging the performance gap to the fine-tuning baselines via two strategies: (1) task-aware demonstration retrieval emphasizes entity and relation information for improving the accuracy of searching demonstrations; (2) gold label-induced reasoning enriches the reasoning evidence of each demonstration. 
To the best of our knowledge, GPT-RE is the first GPT-3 ICL research that significantly outperforms the fine-tuning baseline on three datasets and achieves SOTA on Semeval and SciERC. We implement detailed studies to explore how GPT-3 overcomes the difficulties such as \textsc{null} example influence.


\section*{Limitations}
Despite the overall positive results, GPT-RE still faces two shortcomings: (1) the issue of overpredicting has been significantly alleviated but not completely solved, and the \textsc{null} recall still lags behind full-supervised baselines, especially on the datasets containing a large proportion of \textsc{null} examples such as ACE05 (``95.60\%''); 
(2) Though the task-aware retriever optimizes the representations of PLMs such as SimCSE and BERT, it is widely considered that LLMs can generate more robust representations than small PLMs.
Future work can replace representations generated by smaller PLMs with GPT-3 itself. However, due to the access limitation to the representations of GPT-3, we can nearly confirm this proposal up to now.


\bibliography{anthology,custom}

\begin{thebibliography}{41}
\expandafter\ifx\csname natexlab\endcsname\relax\def\natexlab#1{#1}\fi

\bibitem[{Baldini~Soares et~al.(2019)Baldini~Soares, FitzGerald, Ling, and Kwiatkowski}]{baldini-soares-etal-2019-matching}
Livio Baldini~Soares, Nicholas FitzGerald, Jeffrey Ling, and Tom Kwiatkowski. 2019.
\newblock \href {https://doi.org/10.18653/v1/P19-1279} {Matching the blanks: Distributional similarity for relation learning}.
\newblock In \emph{Proceedings of the 57th Annual Meeting of the Association for Computational Linguistics}, pages 2895--2905, Florence, Italy. Association for Computational Linguistics.

\bibitem[{Banko and Etzioni(2008)}]{banko-etzioni-2008-tradeoffs}
Michele Banko and Oren Etzioni. 2008.
\newblock \href {https://aclanthology.org/P08-1004} {The tradeoffs between open and traditional relation extraction}.
\newblock In \emph{Proceedings of ACL-08: HLT}, pages 28--36, Columbus, Ohio. Association for Computational Linguistics.

\bibitem[{Beltagy et~al.(2019)Beltagy, Lo, and Cohan}]{beltagy-etal-2019-scibert}
Iz~Beltagy, Kyle Lo, and Arman Cohan. 2019.
\newblock \href {https://doi.org/10.18653/v1/D19-1371} {{S}ci{BERT}: A pretrained language model for scientific text}.
\newblock In \emph{Proceedings of the 2019 Conference on Empirical Methods in Natural Language Processing and the 9th International Joint Conference on Natural Language Processing (EMNLP-IJCNLP)}, pages 3615--3620, Hong Kong, China. Association for Computational Linguistics.

\bibitem[{Blevins et~al.(2022)Blevins, Gonen, and Zettlemoyer}]{DBLP:journals/corr/abs-2211-07830}
Terra Blevins, Hila Gonen, and Luke Zettlemoyer. 2022.
\newblock \href {https://doi.org/10.48550/arXiv.2211.07830} {Prompting language models for linguistic structure}.
\newblock \emph{CoRR}, abs/2211.07830.

\bibitem[{Bowman et~al.(2015)Bowman, Angeli, Potts, and Manning}]{bowman-etal-2015-large}
Samuel~R. Bowman, Gabor Angeli, Christopher Potts, and Christopher~D. Manning. 2015.
\newblock \href {https://doi.org/10.18653/v1/D15-1075} {A large annotated corpus for learning natural language inference}.
\newblock In \emph{Proceedings of the 2015 Conference on Empirical Methods in Natural Language Processing}, pages 632--642, Lisbon, Portugal. Association for Computational Linguistics.

\bibitem[{Brown et~al.(2020)Brown, Mann, Ryder, Subbiah, Kaplan, Dhariwal, Neelakantan, Shyam, Sastry, Askell, Agarwal, Herbert-Voss, Krueger, Henighan, Child, Ramesh, Ziegler, Wu, Winter, Hesse, Chen, Sigler, Litwin, Gray, Chess, Clark, Berner, McCandlish, Radford, Sutskever, and Amodei}]{NEURIPS2020_1457c0d6}
Tom Brown, Benjamin Mann, Nick Ryder, Melanie Subbiah, Jared~D Kaplan, Prafulla Dhariwal, Arvind Neelakantan, Pranav Shyam, Girish Sastry, Amanda Askell, Sandhini Agarwal, Ariel Herbert-Voss, Gretchen Krueger, Tom Henighan, Rewon Child, Aditya Ramesh, Daniel Ziegler, Jeffrey Wu, Clemens Winter, Chris Hesse, Mark Chen, Eric Sigler, Mateusz Litwin, Scott Gray, Benjamin Chess, Jack Clark, Christopher Berner, Sam McCandlish, Alec Radford, Ilya Sutskever, and Dario Amodei. 2020.
\newblock \href {https://proceedings.neurips.cc/paper/2020/file/1457c0d6bfcb4967418bfb8ac142f64a-Paper.pdf} {Language models are few-shot learners}.
\newblock In \emph{Advances in Neural Information Processing Systems}, volume~33, pages 1877--1901. Curran Associates, Inc.

\bibitem[{Chowdhery et~al.(2022)Chowdhery, Narang, Devlin, Bosma, Mishra, Roberts, Barham, Chung, Sutton, Gehrmann, Schuh, Shi, Tsvyashchenko, Maynez, Rao, Barnes, Tay, Shazeer, Prabhakaran, Reif, Du, Hutchinson, Pope, Bradbury, Austin, Isard, Gur{-}Ari, Yin, Duke, Levskaya, Ghemawat, Dev, Michalewski, Garcia, Misra, Robinson, Fedus, Zhou, Ippolito, Luan, Lim, Zoph, Spiridonov, Sepassi, Dohan, Agrawal, Omernick, Dai, Pillai, Pellat, Lewkowycz, Moreira, Child, Polozov, Lee, Zhou, Wang, Saeta, Diaz, Firat, Catasta, Wei, Meier{-}Hellstern, Eck, Dean, Petrov, and Fiedel}]{DBLP:journals/corr/abs-2204-02311}
Aakanksha Chowdhery, Sharan Narang, Jacob Devlin, Maarten Bosma, Gaurav Mishra, Adam Roberts, Paul Barham, Hyung~Won Chung, Charles Sutton, Sebastian Gehrmann, Parker Schuh, Kensen Shi, Sasha Tsvyashchenko, Joshua Maynez, Abhishek Rao, Parker Barnes, Yi~Tay, Noam Shazeer, Vinodkumar Prabhakaran, Emily Reif, Nan Du, Ben Hutchinson, Reiner Pope, James Bradbury, Jacob Austin, Michael Isard, Guy Gur{-}Ari, Pengcheng Yin, Toju Duke, Anselm Levskaya, Sanjay Ghemawat, Sunipa Dev, Henryk Michalewski, Xavier Garcia, Vedant Misra, Kevin Robinson, Liam Fedus, Denny Zhou, Daphne Ippolito, David Luan, Hyeontaek Lim, Barret Zoph, Alexander Spiridonov, Ryan Sepassi, David Dohan, Shivani Agrawal, Mark Omernick, Andrew~M. Dai, Thanumalayan~Sankaranarayana Pillai, Marie Pellat, Aitor Lewkowycz, Erica Moreira, Rewon Child, Oleksandr Polozov, Katherine Lee, Zongwei Zhou, Xuezhi Wang, Brennan Saeta, Mark Diaz, Orhan Firat, Michele Catasta, Jason Wei, Kathy Meier{-}Hellstern, Douglas Eck, Jeff Dean, Slav Petrov, and Noah Fiedel.
  2022.
\newblock \href {https://doi.org/10.48550/arXiv.2204.02311} {Palm: Scaling language modeling with pathways}.
\newblock \emph{CoRR}, abs/2204.02311.

\bibitem[{Cohen et~al.(2020)Cohen, Rosenman, and Goldberg}]{DBLP:journals/corr/abs-2010-04829}
Amir D.~N. Cohen, Shachar Rosenman, and Yoav Goldberg. 2020.
\newblock \href {http://arxiv.org/abs/2010.04829} {Relation extraction as two-way span-prediction}.
\newblock \emph{CoRR}, abs/2010.04829.

\bibitem[{Devlin et~al.(2019)Devlin, Chang, Lee, and Toutanova}]{devlin-etal-2019-bert}
Jacob Devlin, Ming-Wei Chang, Kenton Lee, and Kristina Toutanova. 2019.
\newblock \href {https://doi.org/10.18653/v1/N19-1423} {{BERT}: Pre-training of deep bidirectional transformers for language understanding}.
\newblock In \emph{Proceedings of the 2019 Conference of the North {A}merican Chapter of the Association for Computational Linguistics: Human Language Technologies, Volume 1 (Long and Short Papers)}, pages 4171--4186, Minneapolis, Minnesota. Association for Computational Linguistics.

\bibitem[{Gao et~al.(2021)Gao, Yao, and Chen}]{gao-etal-2021-simcse}
Tianyu Gao, Xingcheng Yao, and Danqi Chen. 2021.
\newblock \href {https://doi.org/10.18653/v1/2021.emnlp-main.552} {{S}im{CSE}: Simple contrastive learning of sentence embeddings}.
\newblock In \emph{Proceedings of the 2021 Conference on Empirical Methods in Natural Language Processing}, pages 6894--6910, Online and Punta Cana, Dominican Republic. Association for Computational Linguistics.

\bibitem[{Geng et~al.(2020)Geng, Chen, Zhu, Shen, and Zhao}]{DBLP:conf/cikm/GengCZSZ20}
Xiaoqing Geng, Xiwen Chen, Kenny~Q. Zhu, Libin Shen, and Yinggong Zhao. 2020.
\newblock \href {https://doi.org/10.1145/3340531.3411858} {{MICK:} {A} meta-learning framework for few-shot relation classification with small training data}.
\newblock In \emph{{CIKM} '20: The 29th {ACM} International Conference on Information and Knowledge Management, Virtual Event, Ireland, October 19-23, 2020}, pages 415--424. {ACM}.

\bibitem[{Guti{\'{e}}rrez et~al.(2022)Guti{\'{e}}rrez, McNeal, Washington, Chen, Li, Sun, and Su}]{DBLP:journals/corr/abs-2203-08410}
Bernal~Jim{\'{e}}nez Guti{\'{e}}rrez, Nikolas McNeal, Clay Washington, You Chen, Lang Li, Huan Sun, and Yu~Su. 2022.
\newblock \href {https://doi.org/10.48550/arXiv.2203.08410} {Thinking about {GPT-3} in-context learning for biomedical ie? think again}.
\newblock \emph{CoRR}, abs/2203.08410.

\bibitem[{Han et~al.(2018)Han, Zhu, Yu, Wang, Yao, Liu, and Sun}]{han-etal-2018-fewrel}
Xu~Han, Hao Zhu, Pengfei Yu, Ziyun Wang, Yuan Yao, Zhiyuan Liu, and Maosong Sun. 2018.
\newblock \href {https://doi.org/10.18653/v1/D18-1514} {{F}ew{R}el: A large-scale supervised few-shot relation classification dataset with state-of-the-art evaluation}.
\newblock In \emph{Proceedings of the 2018 Conference on Empirical Methods in Natural Language Processing}, pages 4803--4809, Brussels, Belgium. Association for Computational Linguistics.

\bibitem[{Hendrickx et~al.(2010)Hendrickx, Kim, Kozareva, Nakov, {\'O}~S{\'e}aghdha, Pad{\'o}, Pennacchiotti, Romano, and Szpakowicz}]{hendrickx-etal-2010-semeval}
Iris Hendrickx, Su~Nam Kim, Zornitsa Kozareva, Preslav Nakov, Diarmuid {\'O}~S{\'e}aghdha, Sebastian Pad{\'o}, Marco Pennacchiotti, Lorenza Romano, and Stan Szpakowicz. 2010.
\newblock \href {https://aclanthology.org/S10-1006} {{S}em{E}val-2010 task 8: Multi-way classification of semantic relations between pairs of nominals}.
\newblock In \emph{Proceedings of the 5th International Workshop on Semantic Evaluation}, pages 33--38, Uppsala, Sweden. Association for Computational Linguistics.

\bibitem[{Hoffmann et~al.(2022)Hoffmann, Borgeaud, Mensch, Buchatskaya, Cai, Rutherford, Casas, Hendricks, Welbl, Clark, Hennigan, Noland, Millican, Driessche, Damoc, Guy, Osindero, Simonyan, Elsen, Rae, Vinyals, and Sifre}]{https://doi.org/10.48550/arxiv.2203.15556}
Jordan Hoffmann, Sebastian Borgeaud, Arthur Mensch, Elena Buchatskaya, Trevor Cai, Eliza Rutherford, Diego de~Las Casas, Lisa~Anne Hendricks, Johannes Welbl, Aidan Clark, Tom Hennigan, Eric Noland, Katie Millican, George van~den Driessche, Bogdan Damoc, Aurelia Guy, Simon Osindero, Karen Simonyan, Erich Elsen, Jack~W. Rae, Oriol Vinyals, and Laurent Sifre. 2022.
\newblock \href {https://doi.org/10.48550/ARXIV.2203.15556} {Training compute-optimal large language models}.

\bibitem[{Kojima et~al.(2022)Kojima, Gu, Reid, Matsuo, and Iwasawa}]{DBLP:journals/corr/abs-2205-11916}
Takeshi Kojima, Shixiang~Shane Gu, Machel Reid, Yutaka Matsuo, and Yusuke Iwasawa. 2022.
\newblock \href {https://doi.org/10.48550/arXiv.2205.11916} {Large language models are zero-shot reasoners}.
\newblock \emph{CoRR}, abs/2205.11916.

\bibitem[{Lan et~al.(2019)Lan, Chen, Goodman, Gimpel, Sharma, and Soricut}]{https://doi.org/10.48550/arxiv.1909.11942}
Zhenzhong Lan, Mingda Chen, Sebastian Goodman, Kevin Gimpel, Piyush Sharma, and Radu Soricut. 2019.
\newblock \href {https://doi.org/10.48550/ARXIV.1909.11942} {Albert: A lite bert for self-supervised learning of language representations}.

\bibitem[{Liu et~al.(2022{\natexlab{a}})Liu, Lin, Han, Cao, and Sun}]{liu-etal-2022-pre}
Fangchao Liu, Hongyu Lin, Xianpei Han, Boxi Cao, and Le~Sun. 2022{\natexlab{a}}.
\newblock \href {https://doi.org/10.18653/v1/2022.acl-long.397} {Pre-training to match for unified low-shot relation extraction}.
\newblock In \emph{Proceedings of the 60th Annual Meeting of the Association for Computational Linguistics (Volume 1: Long Papers)}, pages 5785--5795, Dublin, Ireland. Association for Computational Linguistics.

\bibitem[{Liu et~al.(2022{\natexlab{b}})Liu, Shen, Zhang, Dolan, Carin, and Chen}]{liu-etal-2022-makes}
Jiachang Liu, Dinghan Shen, Yizhe Zhang, Bill Dolan, Lawrence Carin, and Weizhu Chen. 2022{\natexlab{b}}.
\newblock \href {https://doi.org/10.18653/v1/2022.deelio-1.10} {What makes good in-context examples for {GPT}-3?}
\newblock In \emph{Proceedings of Deep Learning Inside Out (DeeLIO 2022): The 3rd Workshop on Knowledge Extraction and Integration for Deep Learning Architectures}, pages 100--114, Dublin, Ireland and Online. Association for Computational Linguistics.

\bibitem[{Lu et~al.(2022)Lu, Bartolo, Moore, Riedel, and Stenetorp}]{lu-etal-2022-fantastically}
Yao Lu, Max Bartolo, Alastair Moore, Sebastian Riedel, and Pontus Stenetorp. 2022.
\newblock \href {https://doi.org/10.18653/v1/2022.acl-long.556} {Fantastically ordered prompts and where to find them: Overcoming few-shot prompt order sensitivity}.
\newblock In \emph{Proceedings of the 60th Annual Meeting of the Association for Computational Linguistics (Volume 1: Long Papers)}, pages 8086--8098, Dublin, Ireland. Association for Computational Linguistics.

\bibitem[{Luan et~al.(2018)Luan, He, Ostendorf, and Hajishirzi}]{luan-etal-2018-multi}
Yi~Luan, Luheng He, Mari Ostendorf, and Hannaneh Hajishirzi. 2018.
\newblock \href {https://doi.org/10.18653/v1/D18-1360} {Multi-task identification of entities, relations, and coreference for scientific knowledge graph construction}.
\newblock In \emph{Proceedings of the 2018 Conference on Empirical Methods in Natural Language Processing}, pages 3219--3232, Brussels, Belgium. Association for Computational Linguistics.

\bibitem[{Malkin et~al.(2022)Malkin, Wang, and Jojic}]{malkin-etal-2022-coherence}
Nikolay Malkin, Zhen Wang, and Nebojsa Jojic. 2022.
\newblock \href {https://doi.org/10.18653/v1/2022.acl-long.565} {Coherence boosting: When your pretrained language model is not paying enough attention}.
\newblock In \emph{Proceedings of the 60th Annual Meeting of the Association for Computational Linguistics (Volume 1: Long Papers)}, pages 8214--8236, Dublin, Ireland. Association for Computational Linguistics.

\bibitem[{Min et~al.(2022{\natexlab{a}})Min, Lyu, Holtzman, Artetxe, Lewis, Hajishirzi, and Zettlemoyer}]{https://doi.org/10.48550/arxiv.2202.12837}
Sewon Min, Xinxi Lyu, Ari Holtzman, Mikel Artetxe, Mike Lewis, Hannaneh Hajishirzi, and Luke Zettlemoyer. 2022{\natexlab{a}}.
\newblock \href {https://doi.org/10.48550/ARXIV.2202.12837} {Rethinking the role of demonstrations: What makes in-context learning work?}

\bibitem[{Min et~al.(2022{\natexlab{b}})Min, Lyu, Holtzman, Artetxe, Lewis, Hajishirzi, and Zettlemoyer}]{DBLP:journals/corr/abs-2202-12837}
Sewon Min, Xinxi Lyu, Ari Holtzman, Mikel Artetxe, Mike Lewis, Hannaneh Hajishirzi, and Luke Zettlemoyer. 2022{\natexlab{b}}.
\newblock \href {http://arxiv.org/abs/2202.12837} {Rethinking the role of demonstrations: What makes in-context learning work?}
\newblock \emph{CoRR}, abs/2202.12837.

\bibitem[{Perez et~al.(2021)Perez, Kiela, and Cho}]{DBLP:conf/nips/PerezKC21}
Ethan Perez, Douwe Kiela, and Kyunghyun Cho. 2021.
\newblock \href {https://proceedings.neurips.cc/paper/2021/hash/5c04925674920eb58467fb52ce4ef728-Abstract.html} {True few-shot learning with language models}.
\newblock In \emph{Advances in Neural Information Processing Systems 34: Annual Conference on Neural Information Processing Systems 2021, NeurIPS 2021, December 6-14, 2021, virtual}, pages 11054--11070.

\bibitem[{Rae et~al.(2021)Rae, Borgeaud, Cai, Millican, Hoffmann, Song, Aslanides, Henderson, Ring, Young, Rutherford, Hennigan, Menick, Cassirer, Powell, Driessche, Hendricks, Rauh, Huang, Glaese, Welbl, Dathathri, Huang, Uesato, Mellor, Higgins, Creswell, McAleese, Wu, Elsen, Jayakumar, Buchatskaya, Budden, Sutherland, Simonyan, Paganini, Sifre, Martens, Li, Kuncoro, Nematzadeh, Gribovskaya, Donato, Lazaridou, Mensch, Lespiau, Tsimpoukelli, Grigorev, Fritz, Sottiaux, Pajarskas, Pohlen, Gong, Toyama, d'Autume, Li, Terzi, Mikulik, Babuschkin, Clark, Casas, Guy, Jones, Bradbury, Johnson, Hechtman, Weidinger, Gabriel, Isaac, Lockhart, Osindero, Rimell, Dyer, Vinyals, Ayoub, Stanway, Bennett, Hassabis, Kavukcuoglu, and Irving}]{https://doi.org/10.48550/arxiv.2112.11446}
Jack~W. Rae, Sebastian Borgeaud, Trevor Cai, Katie Millican, Jordan Hoffmann, Francis Song, John Aslanides, Sarah Henderson, Roman Ring, Susannah Young, Eliza Rutherford, Tom Hennigan, Jacob Menick, Albin Cassirer, Richard Powell, George van~den Driessche, Lisa~Anne Hendricks, Maribeth Rauh, Po-Sen Huang, Amelia Glaese, Johannes Welbl, Sumanth Dathathri, Saffron Huang, Jonathan Uesato, John Mellor, Irina Higgins, Antonia Creswell, Nat McAleese, Amy Wu, Erich Elsen, Siddhant Jayakumar, Elena Buchatskaya, David Budden, Esme Sutherland, Karen Simonyan, Michela Paganini, Laurent Sifre, Lena Martens, Xiang~Lorraine Li, Adhiguna Kuncoro, Aida Nematzadeh, Elena Gribovskaya, Domenic Donato, Angeliki Lazaridou, Arthur Mensch, Jean-Baptiste Lespiau, Maria Tsimpoukelli, Nikolai Grigorev, Doug Fritz, Thibault Sottiaux, Mantas Pajarskas, Toby Pohlen, Zhitao Gong, Daniel Toyama, Cyprien de~Masson d'Autume, Yujia Li, Tayfun Terzi, Vladimir Mikulik, Igor Babuschkin, Aidan Clark, Diego de~Las Casas, Aurelia Guy, Chris Jones,
  James Bradbury, Matthew Johnson, Blake Hechtman, Laura Weidinger, Iason Gabriel, William Isaac, Ed~Lockhart, Simon Osindero, Laura Rimell, Chris Dyer, Oriol Vinyals, Kareem Ayoub, Jeff Stanway, Lorrayne Bennett, Demis Hassabis, Koray Kavukcuoglu, and Geoffrey Irving. 2021.
\newblock \href {https://doi.org/10.48550/ARXIV.2112.11446} {Scaling language models: Methods, analysis \&amp; insights from training gopher}.

\bibitem[{Raffel et~al.(2019)Raffel, Shazeer, Roberts, Lee, Narang, Matena, Zhou, Li, and Liu}]{https://doi.org/10.48550/arxiv.1910.10683}
Colin Raffel, Noam Shazeer, Adam Roberts, Katherine Lee, Sharan Narang, Michael Matena, Yanqi Zhou, Wei Li, and Peter~J. Liu. 2019.
\newblock \href {https://doi.org/10.48550/ARXIV.1910.10683} {Exploring the limits of transfer learning with a unified text-to-text transformer}.

\bibitem[{Reimers and Gurevych(2019)}]{reimers-gurevych-2019-sentence}
Nils Reimers and Iryna Gurevych. 2019.
\newblock \href {https://doi.org/10.18653/v1/D19-1410} {Sentence-{BERT}: Sentence embeddings using {S}iamese {BERT}-networks}.
\newblock In \emph{Proceedings of the 2019 Conference on Empirical Methods in Natural Language Processing and the 9th International Joint Conference on Natural Language Processing (EMNLP-IJCNLP)}, pages 3982--3992, Hong Kong, China. Association for Computational Linguistics.

\bibitem[{Rubin et~al.(2022)Rubin, Herzig, and Berant}]{rubin-etal-2022-learning}
Ohad Rubin, Jonathan Herzig, and Jonathan Berant. 2022.
\newblock \href {https://doi.org/10.18653/v1/2022.naacl-main.191} {Learning to retrieve prompts for in-context learning}.
\newblock In \emph{Proceedings of the 2022 Conference of the North American Chapter of the Association for Computational Linguistics: Human Language Technologies}, pages 2655--2671, Seattle, United States. Association for Computational Linguistics.

\bibitem[{Shin et~al.(2021)Shin, Lin, Thomson, Chen, Roy, Platanios, Pauls, Klein, Eisner, and Van~Durme}]{shin-etal-2021-constrained}
Richard Shin, Christopher Lin, Sam Thomson, Charles Chen, Subhro Roy, Emmanouil~Antonios Platanios, Adam Pauls, Dan Klein, Jason Eisner, and Benjamin Van~Durme. 2021.
\newblock \href {https://doi.org/10.18653/v1/2021.emnlp-main.608} {Constrained language models yield few-shot semantic parsers}.
\newblock In \emph{Proceedings of the 2021 Conference on Empirical Methods in Natural Language Processing}, pages 7699--7715, Online and Punta Cana, Dominican Republic. Association for Computational Linguistics.

\bibitem[{Thoppilan et~al.(2022)Thoppilan, De~Freitas, Hall, Shazeer, Kulshreshtha, Cheng, Jin, Bos, Baker, Du, Li, Lee, Zheng, Ghafouri, Menegali, Huang, Krikun, Lepikhin, Qin, Chen, Xu, Chen, Roberts, Bosma, Zhao, Zhou, Chang, Krivokon, Rusch, Pickett, Srinivasan, Man, Meier-Hellstern, Morris, Doshi, Santos, Duke, Soraker, Zevenbergen, Prabhakaran, Diaz, Hutchinson, Olson, Molina, Hoffman-John, Lee, Aroyo, Rajakumar, Butryna, Lamm, Kuzmina, Fenton, Cohen, Bernstein, Kurzweil, Aguera-Arcas, Cui, Croak, Chi, and Le}]{https://doi.org/10.48550/arxiv.2201.08239}
Romal Thoppilan, Daniel De~Freitas, Jamie Hall, Noam Shazeer, Apoorv Kulshreshtha, Heng-Tze Cheng, Alicia Jin, Taylor Bos, Leslie Baker, Yu~Du, YaGuang Li, Hongrae Lee, Huaixiu~Steven Zheng, Amin Ghafouri, Marcelo Menegali, Yanping Huang, Maxim Krikun, Dmitry Lepikhin, James Qin, Dehao Chen, Yuanzhong Xu, Zhifeng Chen, Adam Roberts, Maarten Bosma, Vincent Zhao, Yanqi Zhou, Chung-Ching Chang, Igor Krivokon, Will Rusch, Marc Pickett, Pranesh Srinivasan, Laichee Man, Kathleen Meier-Hellstern, Meredith~Ringel Morris, Tulsee Doshi, Renelito~Delos Santos, Toju Duke, Johnny Soraker, Ben Zevenbergen, Vinodkumar Prabhakaran, Mark Diaz, Ben Hutchinson, Kristen Olson, Alejandra Molina, Erin Hoffman-John, Josh Lee, Lora Aroyo, Ravi Rajakumar, Alena Butryna, Matthew Lamm, Viktoriya Kuzmina, Joe Fenton, Aaron Cohen, Rachel Bernstein, Ray Kurzweil, Blaise Aguera-Arcas, Claire Cui, Marian Croak, Ed~Chi, and Quoc Le. 2022.
\newblock \href {https://doi.org/10.48550/ARXIV.2201.08239} {Lamda: Language models for dialog applications}.

\bibitem[{Wan et~al.(2022)Wan, Liu, Mao, Cheng, Kurohashi, and Li}]{DBLP:journals/corr/abs-2210-11800}
Zhen Wan, Qianying Liu, Zhuoyuan Mao, Fei Cheng, Sadao Kurohashi, and Jiwei Li. 2022.
\newblock \href {https://doi.org/10.48550/arXiv.2210.11800} {Rescue implicit and long-tail cases: Nearest neighbor relation extraction}.
\newblock \emph{CoRR}, abs/2210.11800.

\bibitem[{Wang et~al.(2022{\natexlab{a}})Wang, Liu, Chen, Hong, Tang, and Song}]{wang-etal-2022-deepstruct}
Chenguang Wang, Xiao Liu, Zui Chen, Haoyun Hong, Jie Tang, and Dawn Song. 2022{\natexlab{a}}.
\newblock \href {https://doi.org/10.18653/v1/2022.findings-acl.67} {{D}eep{S}truct: Pretraining of language models for structure prediction}.
\newblock In \emph{Findings of the Association for Computational Linguistics: ACL 2022}, pages 803--823, Dublin, Ireland. Association for Computational Linguistics.

\bibitem[{Wang et~al.(2022{\natexlab{b}})Wang, Wei, Schuurmans, Le, Chi, and Zhou}]{DBLP:journals/corr/abs-2203-11171}
Xuezhi Wang, Jason Wei, Dale Schuurmans, Quoc~V. Le, Ed~H. Chi, and Denny Zhou. 2022{\natexlab{b}}.
\newblock \href {https://doi.org/10.48550/arXiv.2203.11171} {Self-consistency improves chain of thought reasoning in language models}.
\newblock \emph{CoRR}, abs/2203.11171.

\bibitem[{Wei et~al.(2022)Wei, Wang, Schuurmans, Bosma, Chi, Le, and Zhou}]{DBLP:journals/corr/abs-2201-11903}
Jason Wei, Xuezhi Wang, Dale Schuurmans, Maarten Bosma, Ed~H. Chi, Quoc Le, and Denny Zhou. 2022.
\newblock \href {http://arxiv.org/abs/2201.11903} {Chain of thought prompting elicits reasoning in large language models}.
\newblock \emph{CoRR}, abs/2201.11903.

\bibitem[{Williams et~al.(2018)Williams, Nangia, and Bowman}]{williams-etal-2018-broad}
Adina Williams, Nikita Nangia, and Samuel Bowman. 2018.
\newblock \href {https://doi.org/10.18653/v1/N18-1101} {A broad-coverage challenge corpus for sentence understanding through inference}.
\newblock In \emph{Proceedings of the 2018 Conference of the North {A}merican Chapter of the Association for Computational Linguistics: Human Language Technologies, Volume 1 (Long Papers)}, pages 1112--1122, New Orleans, Louisiana. Association for Computational Linguistics.

\bibitem[{Wolf et~al.(2020)Wolf, Debut, Sanh, Chaumond, Delangue, Moi, Cistac, Rault, Louf, Funtowicz, Davison, Shleifer, von Platen, Ma, Jernite, Plu, Xu, Le~Scao, Gugger, Drame, Lhoest, and Rush}]{wolf-etal-2020-transformers}
Thomas Wolf, Lysandre Debut, Victor Sanh, Julien Chaumond, Clement Delangue, Anthony Moi, Pierric Cistac, Tim Rault, Remi Louf, Morgan Funtowicz, Joe Davison, Sam Shleifer, Patrick von Platen, Clara Ma, Yacine Jernite, Julien Plu, Canwen Xu, Teven Le~Scao, Sylvain Gugger, Mariama Drame, Quentin Lhoest, and Alexander Rush. 2020.
\newblock \href {https://doi.org/10.18653/v1/2020.emnlp-demos.6} {Transformers: State-of-the-art natural language processing}.
\newblock In \emph{Proceedings of the 2020 Conference on Empirical Methods in Natural Language Processing: System Demonstrations}, pages 38--45, Online. Association for Computational Linguistics.

\bibitem[{Zhang et~al.(2017)Zhang, Zhong, Chen, Angeli, and Manning}]{zhang-etal-2017-position}
Yuhao Zhang, Victor Zhong, Danqi Chen, Gabor Angeli, and Christopher~D. Manning. 2017.
\newblock \href {https://doi.org/10.18653/v1/D17-1004} {Position-aware attention and supervised data improve slot filling}.
\newblock In \emph{Proceedings of the 2017 Conference on Empirical Methods in Natural Language Processing}, pages 35--45, Copenhagen, Denmark. Association for Computational Linguistics.

\bibitem[{Zhao et~al.(2021)Zhao, Wallace, Feng, Klein, and Singh}]{DBLP:conf/icml/ZhaoWFK021}
Zihao Zhao, Eric Wallace, Shi Feng, Dan Klein, and Sameer Singh. 2021.
\newblock \href {http://proceedings.mlr.press/v139/zhao21c.html} {Calibrate before use: Improving few-shot performance of language models}.
\newblock In \emph{Proceedings of the 38th International Conference on Machine Learning, {ICML} 2021, 18-24 July 2021, Virtual Event}, volume 139 of \emph{Proceedings of Machine Learning Research}, pages 12697--12706. {PMLR}.

\bibitem[{Zhong and Chen(2021)}]{zhong-chen-2021-frustratingly}
Zexuan Zhong and Danqi Chen. 2021.
\newblock \href {https://doi.org/10.18653/v1/2021.naacl-main.5} {A frustratingly easy approach for entity and relation extraction}.
\newblock In \emph{Proceedings of the 2021 Conference of the North American Chapter of the Association for Computational Linguistics: Human Language Technologies}, pages 50--61, Online. Association for Computational Linguistics.

\bibitem[{Zhuang et~al.(2021)Zhuang, Wayne, Ya, and Jun}]{zhuang-etal-2021-robustly}
Liu Zhuang, Lin Wayne, Shi Ya, and Zhao Jun. 2021.
\newblock \href {https://aclanthology.org/2021.ccl-1.108} {A robustly optimized {BERT} pre-training approach with post-training}.
\newblock In \emph{Proceedings of the 20th Chinese National Conference on Computational Linguistics}, pages 1218--1227, Huhhot, China. Chinese Information Processing Society of China.

\end{thebibliography}
\bibliographystyle{acl_natbib}
\clearpage
\appendix

\section{Hyperparameters}
\label{hyperparameters}
\subsection{GPT-3 Hyperparameters}
We use the GPT-3 API during the experiments and set the hyperparameters as in Table~\ref{gpt-3: hyperparameters}. Since the ``Temperature'' is set to be $0.0$, denoting the stable output of GPT-3, we report the result of the single run for all experiments. Due to the input length limitation of GPT-3 and the various average lengths of contexts from each dataset, we set different search ranges for the number of demonstrations of each dataset as shown in Table~\ref{input}.
\subsection{Fine-tuning Baseline PURE}
We follow their \textcolor{black}{single-sentence setup to keep consistency among datasets as Semeval and TACRED are both sentence-level RE datasets}. For the PLMs, we also follow PURE by using \textit{scibert-scivocab-uncased}~\cite{beltagy-etal-2019-scibert} as the base encoder for SciERC and \textit{bert-base-uncased}~\cite{devlin-etal-2019-bert} for the remaining three general domain datasets. We follow hyperparameters in their paper.
We used 2 NVIDIA RTX3090 for training.

\subsection{Sentence Embedding Methods}
~\citet{DBLP:journals/corr/abs-2203-08410} uses the \textsc{\small [CLS]} of RoBERTa-large as the representation in retrieval, ~\citet{liu-etal-2022-makes} fine-tunes RoBERTa-large on two natural language inference (NLI) datasets: SNLI~\cite{bowman-etal-2015-large} and MultiNLI~\cite{williams-etal-2018-broad} to enhance the quality of sentence embedding.
For the sentence embedding method SimCSE in our experiment, we utilize the version: \texttt{sup-simcse-bert-base-uncased}.

\begin{table}[t]
    \centering
    \resizebox{0.6\linewidth}{!}{
    \begin{tabular}{lr}
    \toprule
        Hyperparameter &  In Experiment \\
        \hline
        Engine& text-davinci-003 \\
        Temperature & 0.0\\
        Max\_tokens  &256  \\
        Top\_p  &1  \\
        Frequency\_penalty  &0.0  \\
        Presence\_penalty  &0.0  \\
        Best\_of  &1  \\
        Logprob  &1  \\
        \bottomrule
    \end{tabular}
    }
    \caption{GPT-3 Hyperparamters.}
    \label{gpt-3: hyperparameters}
\end{table}

\begin{table}[t]
    \centering
    \resizebox{0.6\linewidth}{!}{
    \begin{tabular}{lrr}
    \toprule
        Dataset &  Lower bound &  Upper bound \\
        \hline
        Semeval& 5&30\\
        TACRED & 5&15\\
        SciERC  &5&30  \\
        ACE05  &5&25  \\
        \bottomrule
    \end{tabular}
    }
    \caption{Search range for each dataset.}
    \label{input}
\end{table}

\begin{figure}
    \centering
      \begin{subfigure}[b]{0.49\textwidth}
         \centering
         \includegraphics[width=\textwidth]{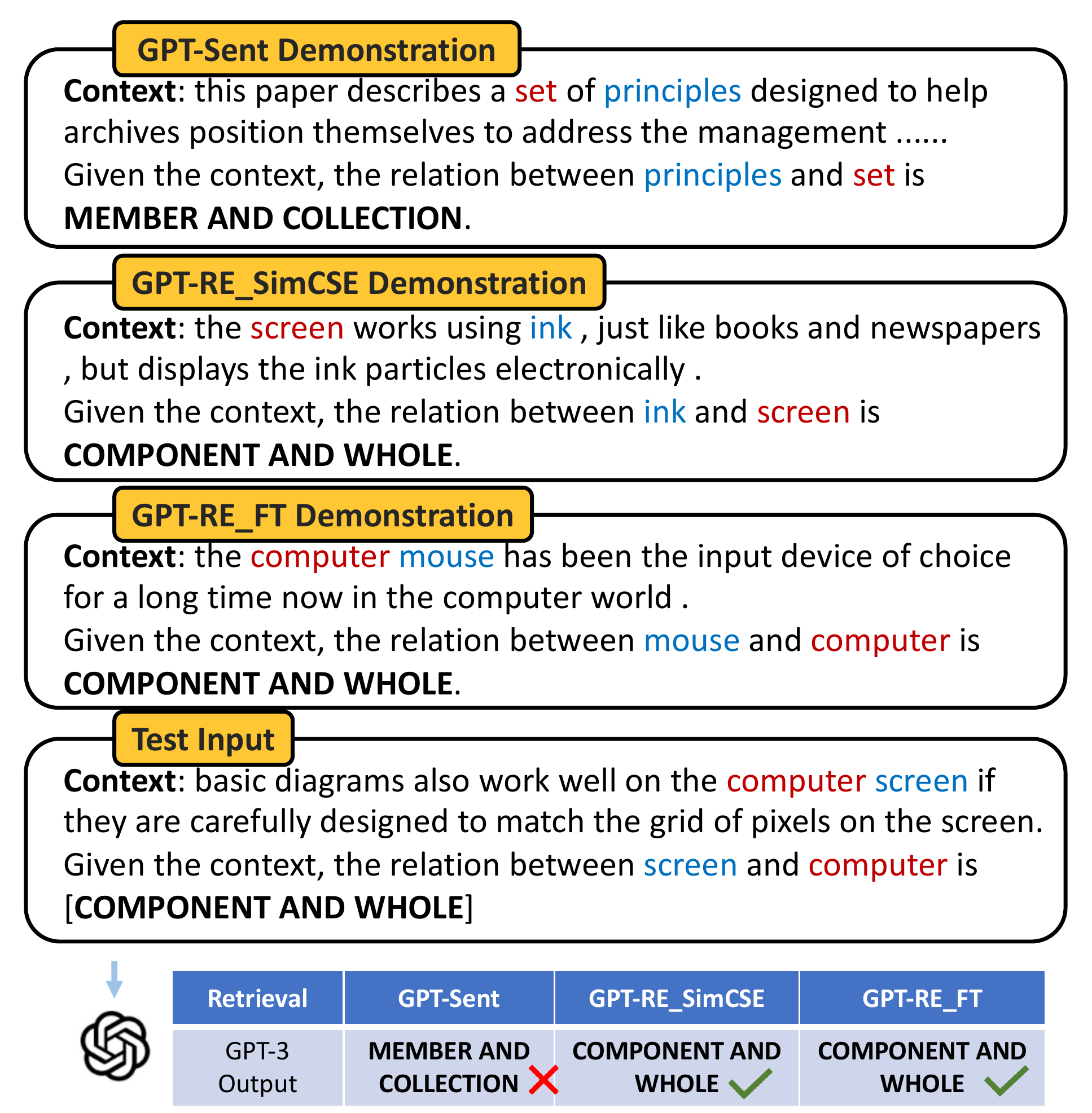}
         \caption{\textbf{[COMPONENT AND WHOLE]} denotes the gold label}
         \label{case1}
     \end{subfigure}
     \begin{subfigure}[b]{0.49\textwidth}
         \centering
         \includegraphics[width=\textwidth]{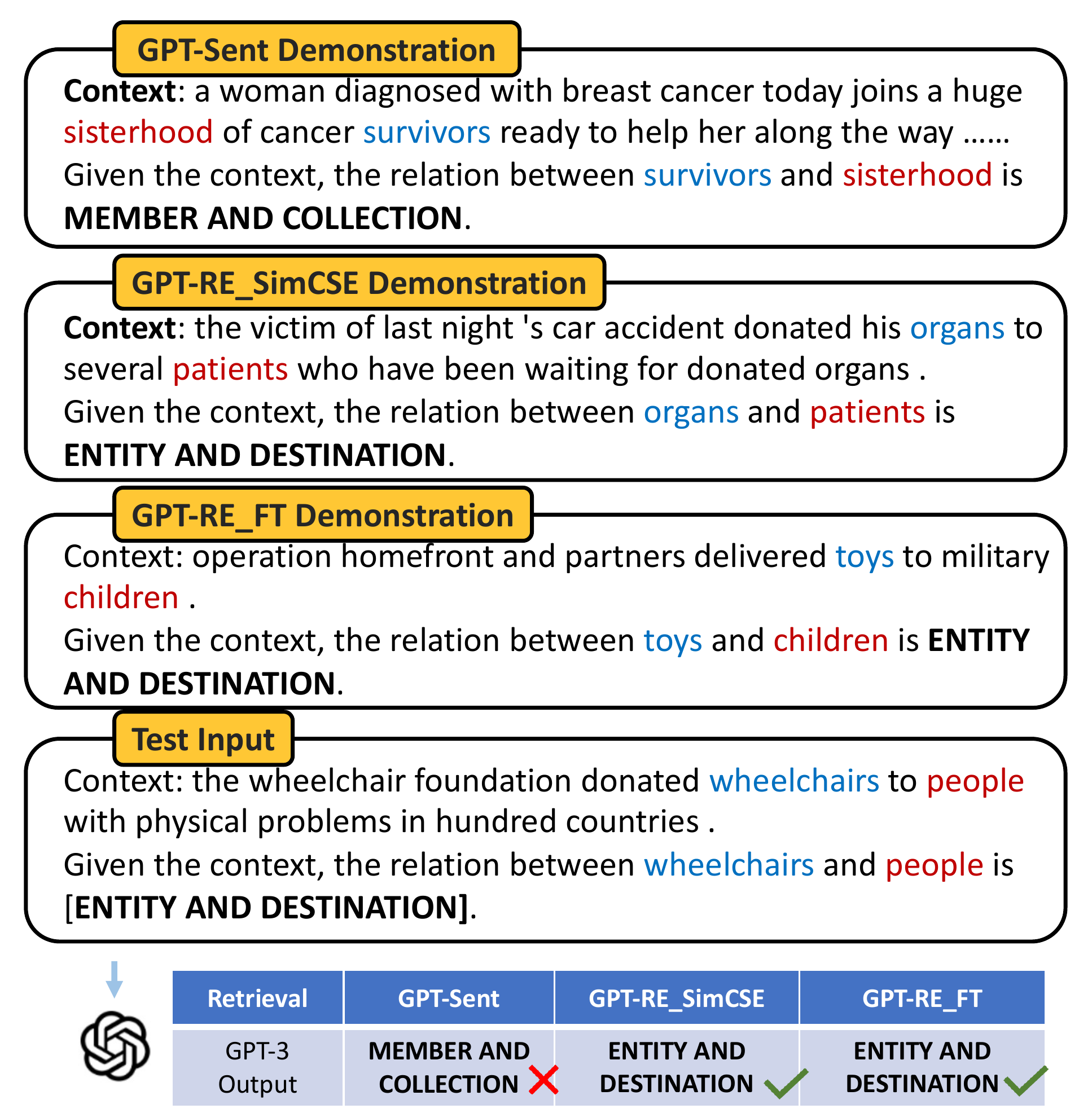}
         \caption{\textbf{[ENTITY AND DESTINATION]} denotes the gold label.}
         \label{case2}
     \end{subfigure}    
    \caption{\textbf{More casees.}}
    
    \label{case12}
\end{figure}

\section{Case Study}
\label{case}
To verify the effectiveness of our task-aware demonstration retrieval, we provide more cases.

For Figure~\ref{case1}, \textit{GPT-Sent} retrieves a demonstration that shares the same semantic meaning of  ``design'' with the test input. However, the entity pair is irrelevant to the concept ``design'' resulting in a noisy demonstration. Instead, \textit{GPT-RE\_SimCSE} retrieves a more relative demonstration with closer pair of entities sharing the same relation label. Furthermore, \textit{GPT-RE\_FT} retrieves the demonstration containing both the closing entity pair and the same linguistic structure between entities. This case emphasizes level-by-level improvement using our proposed methods.
Figure~\ref{case2} shows a similar phenomenon.

\section{Subset}\label{subset}
The number of sampled examples is not only related to the size of the training data itself. A more important factor is the proportion of \textsc{null}. We have to maintain the original label distribution in datasets with a high proportion of \textsc{null}. Thus, the rule to sample the subset is to keep the proportion of each relation label consistent with the original test set. Table~\ref{ace}~\ref{tacred} are label distributions of two subsets.

GPT-RE\_FT on TACRED surpasses the supervised baseline in the current subset. As we show above, some labels in TACRED are indeed not well presented (only $1$ example), since TACRED dataset contains some long-tail labels. We decided to add additional results of GPT-RE\_FT by enlarging our sampled set to \# $3200$ (2 times the current version), and the performance of GPT-RE\_FT (k = $15$) is $73.16$ while the performance of PURE is $70.48$.

\begin{table}[t]
    \centering
    \resizebox{0.6\linewidth}{!}{
    \begin{tabular}{lr}
    \toprule
        Label &  \# Num \\
        \hline
        PHYS& 28\\
        GEN-AFF & 12\\
        PER-SOC  &11  \\
        GEN-AFF  &33  \\
        PART-WHOLE  &13  \\
        ART  &19  \\
        \textsc{null}  &2329  \\
        \bottomrule
    \end{tabular}
    }
    \caption{ACE05}
    \label{ace}
\end{table}

\begin{table}[t]
    \centering
    \resizebox{0.9\linewidth}{!}{
    \begin{tabular}{lr}
    \toprule
        Label &  \# Num \\
        \hline
        Per:title& 40\\
        PER:city\_of\_death & 1\\
        Org:shareholders  &2  \\
        Per:origin  &12  \\
        Org:top\_members/employees	&36  \\
        Org:city\_of\_headquarters &11  \\
        Per:religion	& 4 \\
        Per:city\_of\_birth	& 1\\
        Per:employee\_of	& 27 \\
        Per:data\_of\_death	&3\\
        Per:other\_family	&5\\
        Org:website	&6\\
        Per:cause\_of\_death	&3\\
        Org:subsidiaries	&4\\
        Org:stateorprovince\_of\_headquarters	&5\\
        Per:countries\_of\_residence	&10\\
        Per:siblings	&5\\
        Per:stateorprovinces\_of\_residence	&11\\
        Org:alternate\_names	&27\\
        Per:spouse	&4\\
        Per:parents	&7\\
        Org:country\_of\_headquarters	&9\\
        Per:age	&21\\
        Per:date\_of\_birth	&1\\
        Per:country\_of\_death	&1\\
        Per:schools\_attended	&4\\
        Org:member\_of	&3\\
        Per:children	&5\\
        Org:parents	&7\\
        Per:cities\_of\_residence	&24\\
        Per:stateorprovince\_of\_brith	&1\\
        Per:charges	&12\\
        Org:founded	&2\\
        Org:country\_founded\_by	&5\\
        Per:stateorprovince\_of\_death	&1\\
        Org:members	&4\\
        Per:country\_of\_birth	&1\\
        Per:alternate\_names	&1\\
        Org:number\_of\_employees/members	&1\\
        Org:dissolved	&1\\
        Org:political/religious\_affiliation	&1\\
        \textsc{null}  &1271  \\
        \bottomrule
    \end{tabular}
    }
    \caption{TACRED}
    \label{tacred}
\end{table}

\end{document}